\RequirePackage{fix-cm}
\documentclass{antgroup}
\usepackage[T1]{fontenc}
\usepackage{antgroup}
\usepackage{algorithm}
\usepackage{algpseudocode}
\usepackage{fvextra}
\usepackage{CJKutf8}
\usepackage{needspace}
\usepackage{wrapfig}
\usepackage{float}
\usepackage{pifont}
\usepackage{colortbl}
\usepackage{amsmath,amssymb}
\usepackage{currfile}
\newcommand{\eg}{e.g.}
\usepackage[capitalize,noabbrev]{cleveref}

\title{Describe, Don't Ask: Caption-Driven On-Policy Distillation for Fine-Grained Visual Perception}
\author{\begin{tabular}{@{}c@{}}
Haozhan Shen$^{1,2 *}$\quad Liangbo He$^1$\quad Jun Lan$^{1 \dagger}$\quad Huijia Zhu$^1$\\[3pt]
Weiqiang Wang$^1$\quad Tiancheng Zhao$^{3,4 \dagger}$\quad Jianwei Yin$^2$
\end{tabular}}
\affiliation{\begin{tabular}{@{}c@{}}
$^1$Ant Group\quad $^2$Zhejiang University\\[2pt]
$^3$Om AI Research\quad $^4$Binjiang Institute of Zhejiang University
\end{tabular}}

\begin{document}
\maketitle
\begingroup
\renewcommand{\thefootnote}{}
\makeatletter
\renewcommand{\@makefntext}[1]{\noindent#1}
\makeatother
\begin{NoHyper}
\footnotetext{\textsuperscript{*}Work done during an internship at Ant Group, \textsuperscript{$\dagger$}Corresponding authors\\
Contact: \href{mailto:hz_shen@zju.edu.cn}{hz\_shen@zju.edu.cn}, \href{mailto:yelan.lj@antgroup.com}{yelan.lj@antgroup.com}, \href{mailto:tianchez@zju-bj.com}{tianchez@zju-bj.com}.}
\end{NoHyper}
\endgroup

\begin{abstract}
Vision-language models (VLMs) often recognize fine-grained details
more accurately in a relevant crop than in the full image.
Multimodal on-policy distillation (OPD) methods exploit this
by transferring the advantage of a crop-view teacher to a full-image student.
However, their supervision relies on region-specific questions curated through an elaborate synthesis pipeline,
which brings three weaknesses:
(1) every region needs a question--answer pair written exclusively for it,
locking training into costly QA curation that is hard to scale to unlabeled images;
(2) all rollouts of a region answer the same question,
so the teacher supervises only the details it asks about,
leaving other visual elements in the region unsupervised; and
(3) every question presupposes that its target exists,
so the student becomes prone to answering even when the target is absent.
In this paper, we study how to turn unlabeled images into broad visual supervision
through multi-aspect self-description and privileged-view distillation,
and propose \textbf{CaptionOPD}, which replaces tailored questions with a fixed set of image-agnostic caption prompts,
ranging from general descriptions to focused captions of parts, colors, counts, text, and layout.
The rollouts of a region receive different prompts,
so the teacher supervises various aspects of the region.
Moreover, because caption prompts mask the model to describe what is visible instead of querying a specific target,
they avoid presupposing what the region contains.
We further introduce \emph{caption handoff}, which removes the student's reliance on the region box at test time.
Using the model itself as the crop-view teacher, CaptionOPD requires no questions, no answers, and no larger model annotations: only images and region boxes.
Experiments show that, trained on the same images and regions,
CaptionOPD outperforms prior OPD methods across multiple fine-grained and general perception benchmarks.
Our code is available at \href{https://github.com/inclusionAI/CaptionOPD}{https://github.com/inclusionAI/CaptionOPD}~(will be released soon).
\end{abstract}

\section{Introduction}
\label{sec:intro}

\begin{figure*}[t]
    \centering
    \includegraphics[width=\textwidth]{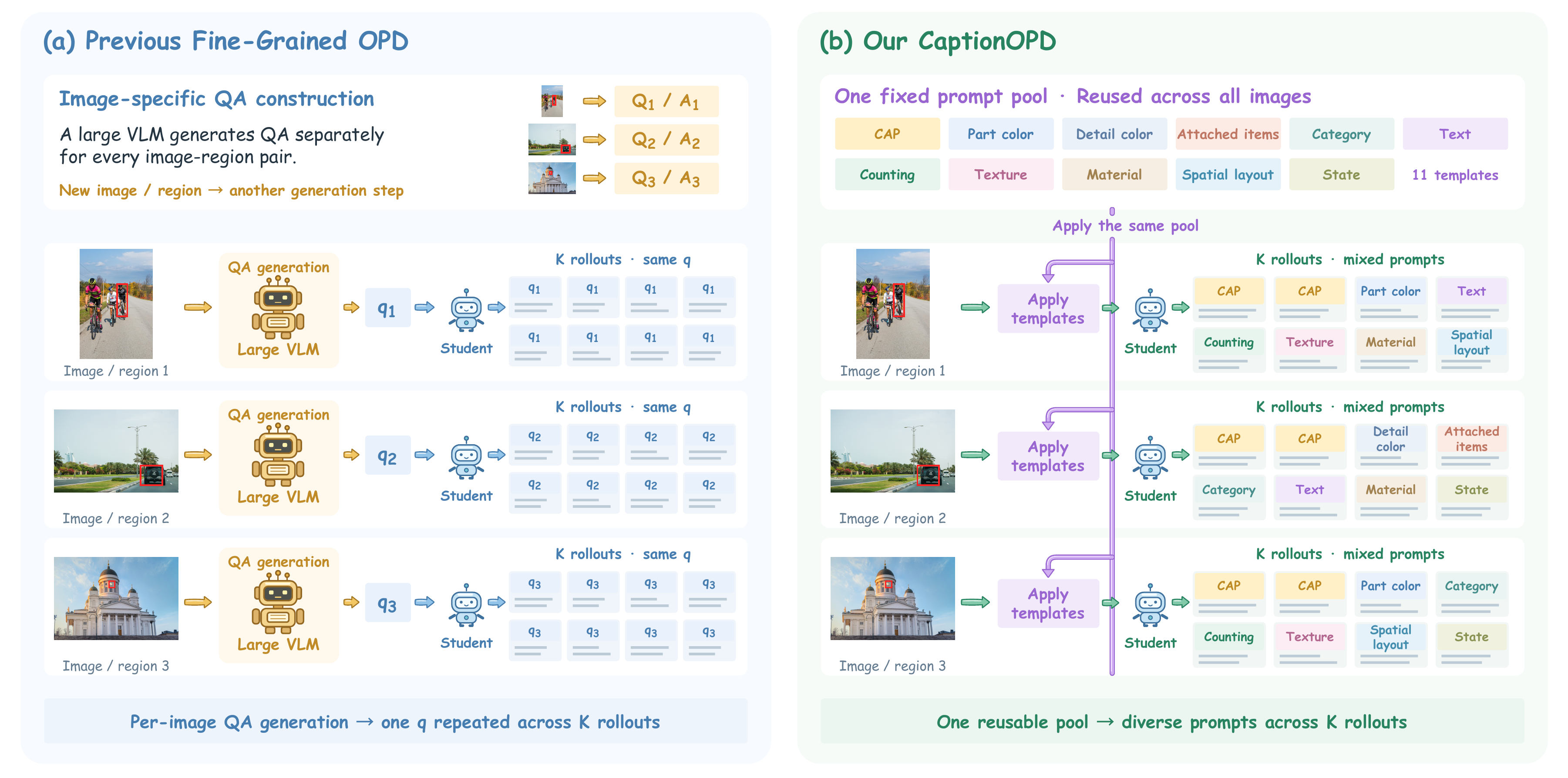}
    \caption{\textbf{Comparison of prompt construction and rollout allocation between previous fine-grained OPD and CaptionOPD.}
        \textbf{(a) Previous fine-grained OPD} relies on a large VLM to synthesize a tailored question $q_i$ for every image--region pair
        and spends all $K$ rollouts answering that same question $q_i$,
        tying training to costly QA curation while supervising only a single queried detail per region.
        \textbf{(b) Our CaptionOPD} eliminates per-region QA generation by reusing a fixed, image-agnostic pool of caption templates
        (detailed in \cref{sec:method-prompts}) across all images,
        assigning diverse prompts across the $K$ rollouts to supervise multiple visual aspects of each region.}
    \label{fig:caption-opd-comparison}
\end{figure*}

Fine-grained visual perception requires Vision-Language Models (VLMs)
to locate and interpret subtle visual evidence in complex scenes,
often involving high-resolution images~(\cite{shen2026zoomgain,wei2026zooming,wu2023vstar,wang2024hrbench,zhang2024mmerealworld,wang2025treebench,lai2025mini-o3}).
While leading VLMs~(\cite{google_gemini3flash,openai2024gpt4ocard,Qwen3-VL,gemmateam2026gemma4technicalreport,qwen35blog})
perform strongly on general visual understanding and reasoning~(\cite{liu2024mmbench,yue2023mmmu,lu2024mathvista,fu2023mme}),
they still struggle with fine-grained perception in complex, high-resolution scenes.
To bridge this gap, several methods enable models to zoom in on relevant regions at test time,
either training-free~(\cite{shen2025zoomeye,zhang2025mllms}),
or by training the model to search and crop regions~(\cite{wu2023vstar,zheng2026deepeyes}).
Yet, these cropping-then-zooming operations adds considerable inference latency.
ZwZ~(\cite{wei2026zooming}) therefore moves zooming into training data construction,
and the model is trained to answer the constructed question on the full image,
so that it perceives details in a single forward pass without tool-call.

Zooming in helps a model locate the key evidence for fine-grained perception
and thus answer questions more accurately.
Motivated by this, Vision-OPD~(\cite{yuan2026visionopd}) adopts the zoomed-in crop as privileged information
and uses a crop-view teacher to supervise a full-image student
via on-policy distillation (OPD)~(\cite{agarwal2024onpolicy}).
To support this training, Vision-OPD uses a large VLM (Qwen3.5-397B-A17B~(\cite{qwen35blog}))
to curate 6K region-specific question--answer pairs.
Follow-up works improve the teacher signal in various ways~(\cite{li2026selfsupervisedvisualonpolicydistillation,opdv2026,zhang2026vad}),
yet they all rely on the exact same training set and QA paradigm, which brings three weaknesses.
\textit{First}, every region requires a large VLM to write a question and its answer, followed by filtering,
which locks training into a costly QA curation pipeline that is hard to scale to massive unlabeled images.
\textit{Second}, the entire rollout budget is spent on the same question (\cref{fig:caption-opd-comparison}a):
it supervises only the details the question asks about,
leaving other visual elements in the region unsupervised.
\textit{Third}, every question presupposes that its target exists
(e.g., asking for the color of a bicycle assumes that a bicycle is in the image).
Thus, the student is prone to committing to a concrete answer even when the queried object is absent.

However, the crop-view teacher does not need a question:
as long as the student describes what it sees in the region,
the teacher can correct it.
Building on this observation,
we study how to turn unlabeled images into broad visual supervision
through multi-aspect self-description and privileged-view distillation,
and propose \textbf{CaptionOPD} (\textbf{Contribution \#1}).
CaptionOPD replaces region-specific questions with a fixed set of image-agnostic caption prompts,
and assigns different prompts to the $K$ rollouts of each region (\cref{fig:caption-opd-comparison}b).
This addresses the three weaknesses in turn:
all regions share the same prompts, so no per-region QA curation is needed;
each rollout describes a different aspect of the region,
so the teacher supervises a variety of visual elements;
and since caption prompts name no target, they do not presuppose that any object exists.
As the teacher is the model itself viewing the zoomed crop,
CaptionOPD requires only images and region boxes.

Unlike a question, a general caption prompt cannot name its target in text,
so the region is marked by a red box,
where the student may come to rely although test images carry no box.
Since the first sentence of a general caption usually names the target,
we propose \textbf{caption handoff} (\textbf{Contribution \#2}):
the caption is generated on the boxed image,
whereas the student is trained on the unmarked image,
with the first sentence serving as a textual cue excluded from the loss (\cref{fig:caption-handoff}).
The student thus learns to find the target and perceive its details on the no-bbox image.

We train Qwen3.5-4B and Qwen3.5-9B with CaptionOPD
on the same images, regions, and rollout budget as prior OPD methods,
using only the images and region boxes (\textbf{Contribution \#3}).
On eight fine-grained perception benchmarks,
CaptionOPD achieves the best average score at both model sizes,
outperforming other OPD baselines.
Furthermore, our case analysis in \cref{sec:presupposition} further shows that
CaptionOPD is less misled by questions that presuppose absent objects than QA-based OPD methods.
Ablations confirm the effectiveness of both caption handoff and the diverse caption prompts (\cref{sec:ablations}).

\section{Background}
\label{sec:background}  

In this section, we first introduce on-policy distillation (\cref{sec:bg-opd}), 
then describe its application on fine-grained visual perception via privileged image crops (\cref{sec:bg-vision-opd}). 
Finally, we introduce the visual target reconstruction objective we adopt in CaptionOPD (\cref{sec:bg-vad}).

\subsection{On-Policy Distillation}
\label{sec:bg-opd} 

Classic knowledge distillation trains a student $p_\theta$ to match a teacher $p_{\bar\theta}$.
However, training on teacher-generated prefixes mismatches the student's own
prefixes during inference.
On-policy distillation (OPD)~(\cite{agarwal2024onpolicy}) mitigates this mismatch by sampling
$y=(y_1,\ldots,y_L)\sim p_\theta(\cdot\mid x)$ for a training input $x\sim\mathcal D$ 
and querying the teacher with the student-generated prefixes.
Writing $p_t^{\mathrm S}=p_\theta(\cdot\mid x,y_{<t})$ and
$p_t^{\mathrm T}=p_{\bar\theta}(\cdot\mid x,y_{<t})$, the objective is
\begin{equation}
    \mathcal L_{\mathrm{OPD}}(\theta)
    =\mathop{\mathbb E}\limits_{\substack{x\sim\mathcal D\\
    y\sim p_\theta(\cdot\mid x)}}
    \left[\frac{1}{L}\sum_{t=1}^{L}
    D\!\left(\operatorname{sg}(p_t^{\mathrm T}),p_t^{\mathrm S}\right)\right],
    \label{eq:opd}
\end{equation}
where $\mathcal D$ is the training data distribution, $D$ is a distribution divergence,
and $\operatorname{sg}$ denotes stop-gradient.

\subsection{Vision-OPD: Distillation across Visual Views}
\label{sec:bg-vision-opd}

Vision-OPD~(\cite{yuan2026visionopd}) applies OPD to transfer a VLM's stronger perception of fine details
in evidence-relevant crops to its full-image policy.
The student and teacher share the same initial model weights and differ in visual input:
given a question $q$, the student sees the full image $I$ with the evidence region marked by a red box,
while the teacher sees the zoomed crop $I^+$ where the details locate.
At the same student-generated prefix, their distributions are
\begin{equation}
    \begin{aligned}
        p_t^{\mathrm S}&=p_\theta(\cdot\mid I,q,y_{<t}),\\
        p_t^{\mathrm T}&=p_{\bar\theta}(\cdot\mid I^+,q,y_{<t}).
    \end{aligned}
    \label{eq:visual-policies}
\end{equation}
With $y\sim p_\theta(\cdot\mid I,q)$, Vision-OPD instantiates $D$ with
Jensen-Shannon divergence (JSD)~(\cite{jsd}):
\begin{equation}
    \begin{aligned}
        &\mathcal L_{\mathrm{Vision\text{-}OPD}}(\theta)\\
        &\quad=\mathop{\mathbb E}\limits_{\substack{(I,I^+,q)\sim\mathcal D\\
        y\sim p_\theta(\cdot\mid I,q)}}
        \left[\frac{1}{L}\sum_{t=1}^{L}
        D_{\mathrm{JS}}\!\left(\operatorname{sg}(p_t^{\mathrm T}),p_t^{\mathrm S}\right)\right].
    \end{aligned}
    \label{eq:vision-opd} 
\end{equation}
Here $D_{\mathrm{JS}}(a,b)=\tfrac12D_{\mathrm{KL}}(a\|m)+\tfrac12D_{\mathrm{KL}}(b\|m)$,
with $m=(a+b)/2$, corresponding to equal mixture weights.
The crop teacher provides token-level supervision during training;
inference adopts only the full-image student.

\subsection{Visual Attribution Distillation}
\label{sec:bg-vad}

Vision-OPD matches the crop teacher’s full distribution,
transferring signals that reflect not only clearer local visual evidence
but may also include effects from linguistic priors and context changes.
Visual Attribution Distillation (VAD)~(\cite{zhang2026vad}) refines it by comparing
the teacher's predictions on clear and degraded crops to construct a target
that emphasizes visually supported signals.

The same teacher evaluates a degraded crop $I^-$ with the question and student prefix unchanged,
giving $p_t^-=p_{\bar\theta}(\cdot\mid I^-,q,y_{<t})$.
Let $\phi(p)=\log(p+\epsilon)-\operatorname{mean}(\log(p+\epsilon))\mathbf 1$
denote centered log probabilities, where $\epsilon>0$ is for numerical stability.
The teacher-student difference $r_t$ and visual evidence direction $u_t$ are
\begin{equation}
    \begin{aligned}
        r_t&=\phi(p_t^{\mathrm T})-\phi(p_t^{\mathrm S}),\\
        u_t&=\phi(p_t^{\mathrm T})-\phi(p_t^-).
    \end{aligned}
    \label{eq:visual-correction}
\end{equation}
VAD uses \(u_t\) to estimate the effect of visual evidence and 
retains the part of the teacher–student difference caused by clearer visual evidence.
\begin{equation}
    \beta_t=\frac{[\langle r_t,u_t\rangle]_+}{\|u_t\|_2^2+\zeta},
    \qquad r_t^{\mathrm{vis}}=\beta_tu_t,
    \label{eq:vad-projection}
\end{equation}
where $[a]_+=\max(a,0)$ and $\zeta>0$ stabilizes the projection.
It then uses \(B_t=\|r_t^{\mathrm{vis}}\|_2\) as the adjustment budget and 
distributes it between support and refutation to yield \(\delta_t\) (detailed in the Appendix~\ref{sec:app-vad}).
The target is reconstructed around the student distribution:
\begin{equation}
    \widehat p_t=\operatorname{softmax}\!\left(
    \phi(p_t^{\mathrm S})+\operatorname{clip}(\delta_t,-c,c)\right),
    \label{eq:vad-target}
\end{equation}
where $c>0$ bounds the coordinate shifts.
VAD matches $\widehat p_t$ using JSD and adds a weak crop-teacher regularizer:
\begin{equation}
    \begin{aligned}
        \ell_t^{\mathrm{VAD}}
        &=D_{\mathrm{JS}}\!\left(\operatorname{sg}(\widehat p_t),p_t^{\mathrm S}\right)\\
        &\quad+\lambda a_tD_{\mathrm{JS}}\!\left(\operatorname{sg}(p_t^{\mathrm T}),p_t^{\mathrm S}\right),\\[3pt]
        \mathcal L_{\mathrm{VAD}}(\theta)
        &=\mathop{\mathbb E}\limits_{\substack{(I,I^+,q)\sim\mathcal D\\
        y\sim p_\theta(\cdot\mid I,q)}}
        \left[\frac{1}{L}\sum_{t=1}^{L}\ell_t^{\mathrm{VAD}}\right],
    \end{aligned}
    \label{eq:vad-objective}
\end{equation}
Here $\lambda$ controls the regularizer. With
$\rho_t=\|\delta_t\|_2/(\|r_t\|_2+\epsilon)$, its token weight is
${a_t=\operatorname{sg}(\operatorname{clip}(1-\rho_t,0,1))}$.
We adopt this target reconstruction and token-level loss for CaptionOPD.

\section{CaptionOPD}
\label{sec:method}

CaptionOPD replaces region-specific questions with fixed caption prompts for crop-to-full-image distillation. 
In this section, we first introduce multi-aspect caption prompts that 
assign diverse description tasks across a region's rollouts as described in \cref{fig:caption-opd-comparison}b (\cref{sec:method-prompts}). 
Next, we present caption handoff, which uses box-guided generations to 
supervise the student on unmarked full images (\cref{sec:method-handoff}). 
Finally, we summarize the overall training procedure (\cref{sec:method-training}).

\begin{figure*}[t]
    \centering
    \includegraphics[width=\textwidth]{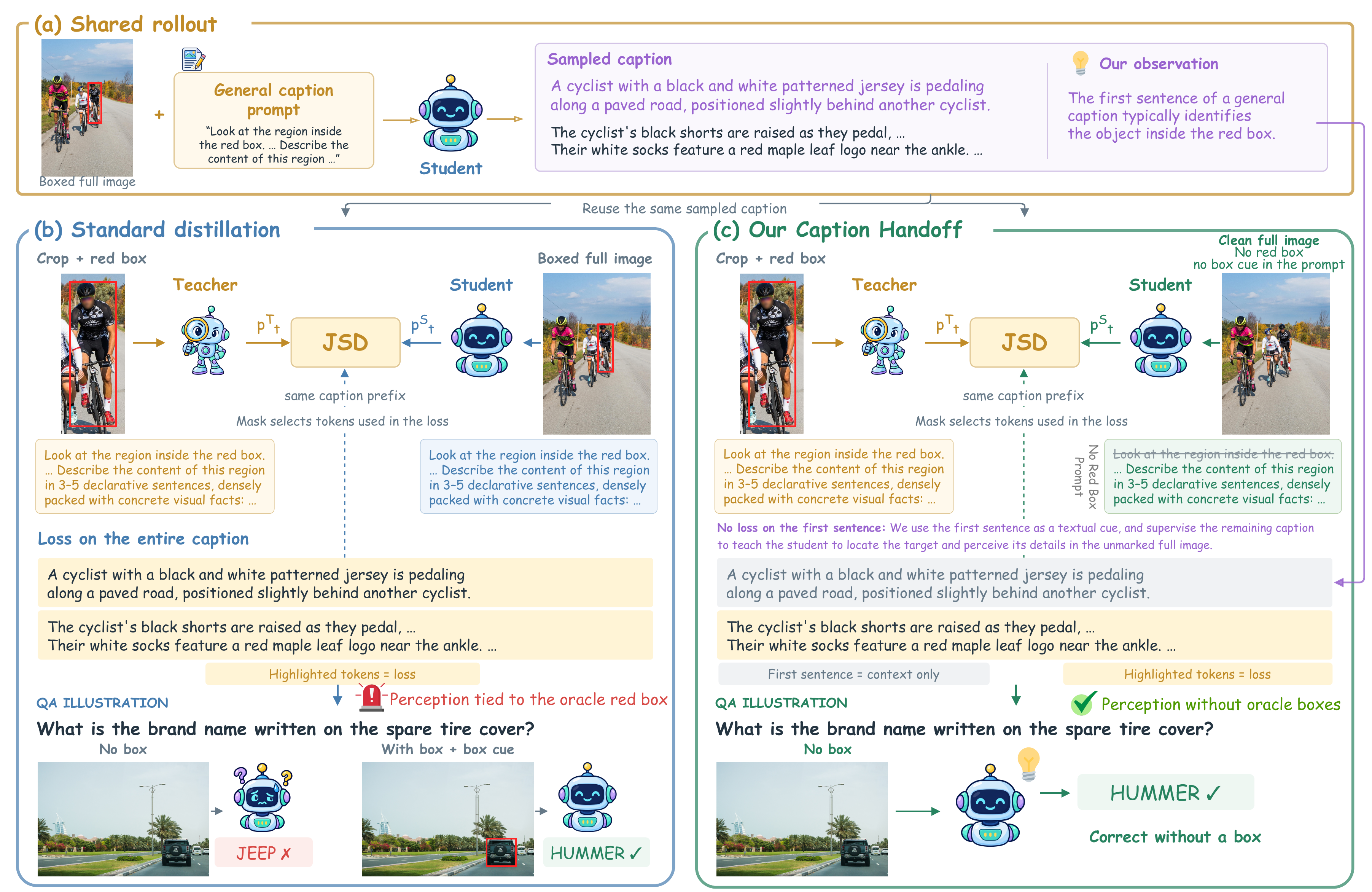}
    \caption{\textbf{Overview of Caption Handoff.} 
    (a) Given a boxed full image, the student generates a general caption 
    whose first sentence typically identifies the target object. 
    (b) Standard distillation retains the box cues during training 
    and supervises the entire caption, 
    causing the student's perception to rely on oracle boxes. 
    (c) The proposed caption handoff removes the box cues 
    from both the student's image and prompt, 
    while keeping the teacher's cropped view unchanged. 
    The first sentence serves purely as a textual anchor (masked from the loss), 
    and only the remaining detailed description is supervised on the clean full image. 
    }
    \label{fig:caption-handoff}
\end{figure*}

\subsection{Multi-Aspect Caption Prompts}
\label{sec:method-prompts}

We define a fixed, image-agnostic prompt pool
$\mathcal T=\{\tau_{\mathrm{CAP}}\}\cup\mathcal T_{\mathrm{focused}}$,
containing one general caption prompt and ten focused prompts.
The general caption prompt asks for an overall description of the region,
while focused prompts direct target specific visual aspects,
such as small-part colors, attached items, fine-grained categories,
text, counts, materials, textures, states, and spatial layout.
Crucially, these prompts merely specify \emph{what} to describe
instead of naming image-specific objects,
and require no ground-truth answers.
\Cref{tab:caption-prompts} summarizes representative description tasks.

\begin{table}[t]
    \centering
    \small
    \setlength{\tabcolsep}{4pt}
    \caption{Representative tasks from the caption prompt pool. 
    Prompts are simplified for brevity; see Appendix~\ref{sec:app-prompts} for the exact wording.}
    \label{tab:caption-prompts}
    \renewcommand{\arraystretch}{0.95}
    \setlength{\aboverulesep}{1pt}
    \setlength{\belowrulesep}{1pt}
    \begin{tabular}{@{}p{0.27\columnwidth}p{0.67\columnwidth}@{}}
        \toprule
        \textbf{Aspect} & \textbf{Description task} \\
        \midrule
        General caption & Describe the region and its visible details. \\
        \midrule
        Small-part color & Identify a small detail on a larger part and give its color. \\
        \midrule
        Attached items & Name an item attached to, held by, or resting on the object. \\
        \midrule
        Text & Transcribe visible text, numbers, or symbols exactly. \\  
        \midrule
        Counting & Select a repeated part or item and count it. \\
        \midrule
        Spatial layout & Describe what appears at a specified side of the object. \\
        \bottomrule
    \end{tabular}
\end{table}

For each region $b$, we allocate $K=8$ rollouts (following prior work~(\cite{yuan2026visionopd,zhang2026vad,opdv2026})) 
across general and focused caption prompts.
We rank regions by their area ratio relative to the full image and divide them into three equal-sized groups.
Since larger regions generally contain richer visual content,
we assign more rollouts to the general caption prompt to allow richer descriptions.
Specifically, small, medium, and large regions receive
$c(b)=1$, $2$, and $3$ general-caption rollouts, respectively.
The remaining $K-c(b)$ rollouts are assigned distinct focused prompts.
Each general-caption rollout uses $\tau_{\mathrm{CAP}}$ with an independently sampled response,
while focused templates are sampled without replacement.
This allocation strategy assigns multiple description tasks to each region
while keeping its total rollout budget fixed.
 
For each selected template $\tau$, we form $q_{\mathrm{box}}$
by prefixing $\tau$ with the box instruction
(e.g., \emph{``Look at the region inside the red box.''}).
This augmented prompt serves as $q$ in \cref{eq:visual-policies}.
The student sees the full image with the target region marked by a box,
while the teacher sees the corresponding zoomed crop.
For each rollout, the student generates a response using its assigned prompt,
and the teacher evaluates the same instruction and response prefix under its crop view.
For general-caption rollouts, we further apply \textit{caption handoff}
to supervise the student, described in \cref{sec:method-handoff}.

\begin{algorithm}[t]
    \caption{CaptionOPD training}
    \label{alg:caption-opd}
    \small
    \begin{algorithmic}[1]
        \Require Student $\theta$, prompt pool $\mathcal T$, rollout budget $K$,
        EMA update rate $\mu$
        \State $\bar\theta\gets\theta$
        \For{each batch $\mathcal B$ of image-region pairs}  
            \State $\mathcal L_{\mathrm{sum}}\gets0$
            \Comment{Loss sum}
            \State $N_{\mathrm{tok}}\gets0$
            \Comment{Supervised token count} 
            \For{each $(I,b)\in\mathcal B$}
                \State Construct $I_{\mathrm{box}},I_{\mathrm{clean}},I^+,I^-$
                \State List $\Pi$: $c(b)$ copies of $\tau_{\mathrm{CAP}}$
                (\cref{sec:method-prompts})
                \State Sample $K-c(b)$ distinct focused prompts into $\Pi$
                \For{each prompt $\tau\in\Pi$} 
                    \State $q_{\mathrm{box}}\gets\Call{AddBoxInstruction}{\tau}$
                    \State Sample $y\sim p_\theta(\cdot\mid I_{\mathrm{box}},q_{\mathrm{box}})$
                    \State $L\gets|y|$
                    \If{$\tau=\tau_{\mathrm{CAP}}$} \Comment{Handoff, \cref{sec:method-handoff}}
                        \State $(I^{\mathrm S},q^{\mathrm S})\gets
                        (I_{\mathrm{clean}},\tau_{\mathrm{CAP}})$
                        \State Find first-sentence length $L^{\mathrm{fs}}$
                        \State $\mathbf m\gets
                        (\underbrace{0,\ldots,0}_{L^{\mathrm{fs}}},
                        \underbrace{1,\ldots,1}_{L-L^{\mathrm{fs}}})^\top$
                    \Else \Comment{Focused supervision}
                        \State $(I^{\mathrm S},q^{\mathrm S})\gets
                        (I_{\mathrm{box}},q_{\mathrm{box}})$
                        \State $\mathbf m\gets\mathbf 1_L$
                    \EndIf
                    \State $p_t^{\mathrm S}\gets
                    p_\theta(\cdot\mid I^{\mathrm S},q^{\mathrm S},y_{<t})$
                    \State $p_t^{\mathrm T}\gets
                    p_{\bar\theta}(\cdot\mid I^+,q_{\mathrm{box}},y_{<t})$
                    \State $p_t^-\gets
                    p_{\bar\theta}(\cdot\mid I^-,q_{\mathrm{box}},y_{<t})$
                    \State Compute token-loss vector via \cref{eq:vad-objective}:
                    \Statex \hspace{4.5em}$\boldsymbol{\ell}^{\mathrm{VAD}}
                    =(\ell_1^{\mathrm{VAD}},\ldots,\ell_L^{\mathrm{VAD}})^\top$
                    \State $\mathcal L_{\mathrm{sum}}\gets
                    \mathcal L_{\mathrm{sum}}
                    +\mathbf m^\top\boldsymbol{\ell}^{\mathrm{VAD}}$
                    \State $N_{\mathrm{tok}}\gets
                    N_{\mathrm{tok}}+\sum_{t=1}^{L}m_t$
                \EndFor
            \EndFor
            \State Update $\theta$ to minimize $\mathcal L_{\mathrm{sum}}/N_{\mathrm{tok}}$
            (\cref{eq:caption-opd-objective})
            \State $\bar\theta\gets(1-\mu)\bar\theta+\mu\theta$
        \EndFor
        \State \Return $\theta$
    \end{algorithmic}
\end{algorithm}

\subsection{Caption Handoff}
\label{sec:method-handoff}

Box cues are necessary to specify which region to describe,
but retaining them throughout distillation can cause the student
to rely on privileged localization information,
which is unavailable during inference on clean full images (\cref{fig:caption-handoff}b).
Meanwhile, we observe that the first sentence of a general caption
typically identifies the target via its category and coarse attributes
(\cref{fig:caption-handoff}a).
Motivated by this, we leverage this target-identifying sentence as a textual cue
to replace the explicit box guidance during supervision,
a procedure we term \emph{caption handoff} (\cref{fig:caption-handoff}c).

Let $I_{\mathrm{box}}$ and $I_{\mathrm{clean}}$ denote the boxed and unmarked full images, respectively.
For general-caption rollouts, $q_{\mathrm{box}}$ consists of
$\tau_{\mathrm{CAP}}$ prefixed with the box instruction.
The student first samples a caption
$y\sim p_\theta(\cdot\mid I_{\mathrm{box}},q_{\mathrm{box}})$ (\cref{fig:caption-handoff}a).
When reusing this caption for distillation,
we remove both the red box and the box instruction
from the student's input,
while the teacher retains its boxed crop $I^+$ and prompt $q_{\mathrm{box}}$.
The student and teacher distributions are then
\begin{equation}
    \begin{aligned}
        p_t^{\mathrm S}
        &=p_\theta(\cdot\mid I_{\mathrm{clean}},\tau_{\mathrm{CAP}},y_{<t}),\\
        p_t^{\mathrm T}
        &=p_{\bar\theta}(\cdot\mid I^+,q_{\mathrm{box}},y_{<t}).
    \end{aligned}
    \label{eq:handoff-policies}
\end{equation}
The sampled caption is reused without regeneration,
with both models conditioned on the same prefix $y_{<t}$.

However, without the red box on $I_{\mathrm{clean}}$,
the student has no explicit cue indicating which object to describe 
at the start of the caption.
To resolve this ambiguity, we treat the first sentence
purely as a textual cue in the prefix and mask it out from the loss
(gray block in \cref{fig:caption-handoff}c). 
Appendix~\ref{sec:app-handoff} details how the first sentence is located
and how captions whose first sentence mentions the box are handled.
Let $L^{\mathrm{fs}}$ denote the token length of the first sentence.
For a caption of length $L>L^{\mathrm{fs}}$, its unnormalized loss contribution
is the sum of VAD token losses (\cref{eq:vad-objective})
over the remaining tokens:
\begin{equation}
    \ell_{\mathrm{handoff}}(y)
    =\sum_{t=L^{\mathrm{fs}}+1}^{L}\ell_t^{\mathrm{VAD}},
    \label{eq:handoff-loss}
\end{equation}
where $\ell_t^{\mathrm{VAD}}$ is evaluated using the handoff distributions
in \cref{eq:handoff-policies}.
Normalization is applied after aggregating all rollouts, using the total number
of supervised tokens in the batch (\cref{eq:caption-opd-objective}).
Conditioned on the first sentence,
the student is trained to locate the described target
and perceive its fine-grained details in the unmarked full image
to match the crop teacher's predictions.

Note that we apply caption handoff exclusively to general-caption rollouts,
as focused prompts directly target fine-grained attributes
(e.g., OCR or counting) without a target-identifying first sentence;
focused rollouts therefore retain the standard box-conditioned supervision.

\subsection{Training Procedure}
\label{sec:method-training}

\Cref{alg:caption-opd} summarizes the overall training procedure.
We initialize the student and teacher from the same pretrained model.
For each image-region pair, we construct the boxed full image $I_{\mathrm{box}}$,
the clean full image $I_{\mathrm{clean}}$,
the clear teacher crop $I^+$, and the degraded teacher crop $I^-$.
We allocate $K$ prompts as described in \cref{sec:method-prompts},
and for each prompt the student generates a caption conditioned on $I_{\mathrm{box}}$.
During supervision, general-caption rollouts apply the handoff conditioning
and first-sentence mask from \cref{sec:method-handoff},
while focused rollouts retain the boxed student input and supervise all tokens.
For every rollout, the teacher evaluates token distributions on both $I^+$ and $I^-$
under $q_{\mathrm{box}}$ to construct the VAD targets (\cref{eq:vad-target})
and compute the token-level losses $\ell_t^{\mathrm{VAD}}$ (\cref{eq:vad-objective}). 

To combine both general-caption and focused-prompt rollouts,
let $i$ index the generated captions in a training batch
and let $L_i$ denote their token lengths.
We introduce a supervision mask $m_{i,t}\in\{0,1\}$:
for general captions, the first $L^{\mathrm{fs}}_i$ tokens are masked out,
while all remaining valid tokens are supervised;
for focused captions, all valid response tokens are supervised.
The batch objective averages over all supervised tokens,
rather than averaging the per-rollout losses:
\begin{equation}
    \mathcal L_{\mathrm{CaptionOPD}}(\theta)
    =\frac{\sum_i\sum_{t=1}^{L_i}
    m_{i,t}\,\ell_{i,t}^{\mathrm{VAD}}}
    {\sum_i\sum_{t=1}^{L_i}m_{i,t}},
    \label{eq:caption-opd-objective}
\end{equation}
where $\ell_{i,t}^{\mathrm{VAD}}$ is the token loss in \cref{eq:vad-objective}
under the corresponding student conditioning.
Thus, each supervised token receives equal weight, 
so captions with more supervised tokens contribute more to the final objective.
We update the student with this objective and update the teacher
by an exponential moving average (EMA) of the student parameters.
Teacher predictions and reconstructed targets are held fixed during backpropagation.

\begin{table*}[t]
\centering
\caption{\textbf{Results on fine-grained perception benchmarks.}
\textit{Unlabeled Experience} denotes learning without question--answer annotations, from images and region boxes only.
Best and second-best scores within each model size are \textbf{bold} and \underline{underlined}.}
\label{tab:main_benchmarks}
\begingroup
\newcommand{\qatick}{\textcolor{green!70!black}{\ding{51}}}
\newcommand{\qacross}{\textcolor{black!60}{\ding{55}}}
\setlength{\aboverulesep}{0pt}
\setlength{\belowrulesep}{0pt}
\setlength{\tabcolsep}{4pt} 
\renewcommand{\arraystretch}{1.24}
\fontsize{8}{9.5}\selectfont
\resizebox{\textwidth}{!}{%
\begin{tabular}{@{}c|*{7}{c}|*{7}{c}@{}}
\toprule[0.9pt]
\multirow{3}{*}{\textbf{Benchmark}} & \multicolumn{7}{c|}{\textbf{Qwen3.5-4B}} & \multicolumn{7}{c}{\textbf{Qwen3.5-9B}} \\
\cmidrule(lr){2-8}\cmidrule(lr){9-15}
 & \multirow{2}{*}{\textbf{Base}} & \multirow{2}{*}{\textbf{S2VOPD}} & \textbf{Vision-} & \multirow{2}{*}{\textbf{OPD-V}} & \multirow{2}{*}{\textbf{VA-OPD}} & \multirow{2}{*}{\textbf{VAD}} & \textbf{Caption-} & \multirow{2}{*}{\textbf{Base}} & \multirow{2}{*}{\textbf{S2VOPD}} & \textbf{Vision-} & \multirow{2}{*}{\textbf{OPD-V}} & \multirow{2}{*}{\textbf{VA-OPD}} & \multirow{2}{*}{\textbf{VAD}} & \textbf{Caption-} \\
 &  &  & \textbf{OPD} &  &  &  & \textbf{OPD} &  &  & \textbf{OPD} &  &  &  & \textbf{OPD} \\
\midrule
\textit{Unlabeled Experience} & -- & \qacross & \qacross & \qacross & \qacross & \qacross & \qatick & -- & \qacross & \qacross & \qacross & \qacross & \qacross & \qatick \\
\midrule
ZoomGainBench & 32.51 & 36.73 & 49.98 & 51.55 & 48.91 & \underline{52.22} & \textbf{58.95} & 36.38 & 40.01 & 53.37 & 53.64 & 54.12 & \underline{57.61} & \textbf{66.00} \\
ZoomBench & 48.40 & 53.25 & 59.64 & \textbf{61.18} & 60.47 & \underline{61.07} & 60.47 & 53.96 & 53.85 & \textbf{64.85} & \underline{63.79} & 62.60 & 62.49 & 61.07 \\
V$^*$Bench & 84.29 & 84.82 & 90.58 & 90.58 & \textbf{94.24} & \underline{93.19} & 87.96 & 83.77 & 89.01 & \textbf{94.24} & 90.05 & \underline{93.72} & 92.67 & \textbf{94.24} \\
HR-Bench 4K & 84.38 & \textbf{85.62} & 80.88 & 82.00 & 83.88 & \underline{85.12} & 84.50 & 83.62 & 84.00 & 86.50 & 84.38 & \textbf{87.12} & 86.75 & \underline{86.88} \\
HR-Bench 8K & 79.12 & \textbf{81.50} & 78.62 & 78.50 & 79.00 & \underline{80.88} & 78.62 & 81.38 & 83.50 & \underline{85.50} & 82.38 & 84.12 & 85.12 & \textbf{85.88} \\
MME-RW-Lite & 57.79 & \underline{61.18} & 58.10 & 56.64 & 58.10 & 61.07 & \textbf{61.28} & 58.83 & 61.18 & 60.97 & 59.61 & 61.02 & \underline{62.12} & \textbf{63.42} \\
TreeBench & \underline{47.16} & 46.17 & 43.70 & 44.20 & 44.94 & 45.19 & \textbf{49.38} & 46.91 & 48.15 & 48.89 & 46.42 & 48.40 & \underline{51.60} & \textbf{53.58} \\
VisualProbe & 43.88 & \underline{52.62} & 52.04 & \underline{52.62} & 49.90 & 52.43 & \textbf{57.28} & 43.88 & 52.23 & 55.92 & 53.40 & 56.50 & \underline{57.67} & \textbf{61.55} \\
\midrule
\textbf{Average} & 59.69 & 62.74 & 64.19 & 64.66 & 64.93 & \underline{66.40} & \textbf{67.30} & 61.09 & 63.99 & 68.78 & 66.71 & 68.45 & \underline{69.50} & \textbf{71.58} \\
\bottomrule[0.9pt]
\end{tabular}%
}
\endgroup
\end{table*}

\section{Experiment}

\subsection{Experimental Settings}
\label{sec:exp-settings}

\paragraph{Training Setup.}
We conduct experiments using Qwen3.5-4B and Qwen3.5-9B~(\cite{qwen35blog}) as the backbone models.

\begin{wraptable}{R}{0.5\textwidth}
\centering
\caption{\textbf{Results on broader perception benchmarks} (Qwen3.5-9B).
Best and second-best scores are \textbf{bold} and \underline{underlined}.}
\label{tab:general_benchmarks}
\begingroup
\setlength{\tabcolsep}{2.5pt}
\renewcommand{\arraystretch}{1.15}
\resizebox{\linewidth}{!}{%
\begin{tabular}{@{}l*{7}{c}@{}}
\toprule
\multirow{2}{*}{\textbf{Benchmark}} & \multirow{2}{*}{\textbf{Base}} & \multirow{2}{*}{\textbf{S2VOPD}} & \textbf{Vision-} & \multirow{2}{*}{\textbf{OPD-V}} & \multirow{2}{*}{\textbf{VA-OPD}} & \multirow{2}{*}{\textbf{VAD}} & \textbf{Caption-} \\
 & & & \textbf{OPD} & & & & \textbf{OPD} \\
\midrule
PerceptionBench & \underline{35.90} & 35.87 & 35.57 & 33.80 & 35.83 & 35.57 & \textbf{39.73} \\
MMVP & 67.33 & 68.67 & 67.33 & 60.67 & \underline{70.67} & 69.33 & \textbf{71.33} \\
BLINK & 67.49 & \underline{68.28} & 66.75 & 59.81 & 67.23 & 68.07 & \textbf{68.39} \\
CV-Bench & 87.74 & 88.27 & 87.78 & 84.72 & \underline{88.37} & \textbf{88.81} & 87.59 \\
CharXiv-DQ & 91.22 & \underline{91.55} & 87.92 & 82.23 & 86.92 & \textbf{91.60} & 90.72 \\
CharXiv-RQ & 66.30 & 66.40 & 63.40 & 60.80 & 64.20 & \underline{67.10} & \textbf{68.30} \\
SEED-Bench-2-Plus & 73.69 & 73.12 & \underline{74.57} & 72.16 & 74.35 & 73.74 & \textbf{74.62} \\
MMStar & \textbf{75.87} & \underline{75.33} & 74.47 & 71.80 & 74.87 & \underline{75.33} & \textbf{75.87} \\
\midrule
\textbf{Average} & 70.69 & 70.94 & 69.72 & 65.75 & 70.31 & \underline{71.19} & \textbf{72.07} \\
\bottomrule
\end{tabular}%
}
\endgroup
\end{wraptable}
\noindent For training, we use the 6,241 image-region pairs from Vision-OPD~(\cite{yuan2026visionopd}),
retaining only the images and bounding boxes.
For each region, we allocate $K=8$ prompts by combining our general caption prompt
and focused description templates (\cref{sec:method-prompts}),
yielding 49,928 training instances in total.
Both the student and teacher are initialized from the same pretrained backbone weights,
with the teacher updated via an exponential moving average (EMA) of the student parameters.
We perform full-parameter training for one epoch
with a learning rate of $2\times 10^{-6}$ and a global batch size of 768.

\paragraph{Baselines.}
We compare CaptionOPD with five recent OPD methods:
Vision-OPD~(\cite{yuan2026visionopd}), S$^2$VOPD~(\cite{li2026selfsupervisedvisualonpolicydistillation}),
VAD~(\cite{zhang2026vad}), VA-OPD~(\cite{liu2026visualadvantageonpolicydistillationvisionlanguage}),
and OPD-V~(\cite{opdv2026}), across both the Qwen3.5 4B and 9B models.
All baseline distillation methods are trained on the exact same Vision-OPD-6K data.
For each image-region sample, the baselines generate 8 rollouts using its original question.

\paragraph{Benchmarks.}
For \emph{fine-grained visual perception}, we evaluate both model sizes on eight benchmarks:
ZoomGainBench~(\cite{shen2026zoomgain}),
ZoomBench~(\cite{wei2026zooming}), V*Bench~(\cite{wu2023vstar}),
HR-Bench 4K/8K~(\cite{wang2024hrbench}), MME-RW-Lite~(\cite{zhang2024mmerealworld}),
TreeBench~(\cite{wang2025treebench}), and VisualProbe~(\cite{lai2025mini-o3}).
To examine \emph{generalization to broader visual tasks},
we further evaluate the 9B models on PerceptionBench~(\cite{lin2026perceptionbench}), MMVP~(\cite{tong2024eyes}),
BLINK~(\cite{fu2024blink}), CV-Bench~(\cite{tong2024cambrian1}),
CharXiv-DQ/RQ~(\cite{wang2024charxiv}), SEED-Bench-2-Plus~(\cite{li2024seed2plus}),
and MMStar~(\cite{chen2024mmstar}).

\paragraph{Evaluation Protocol.}
All models are evaluated with identical prompts and decoding configurations
on each benchmark: disable thinking mode and use greedy decoding
with a maximum output length of 4,096 tokens.
For benchmarks requiring LLM-based judging,
we uniformly employ Qwen3.8-27B~(\cite{qwen3.8}) as the judge.
The prompts of all benchmarks and judges are provided in Appendix~\ref{sec:app-eval}.

\begin{figure*}[t]
    \centering
    \includegraphics[width=\textwidth]{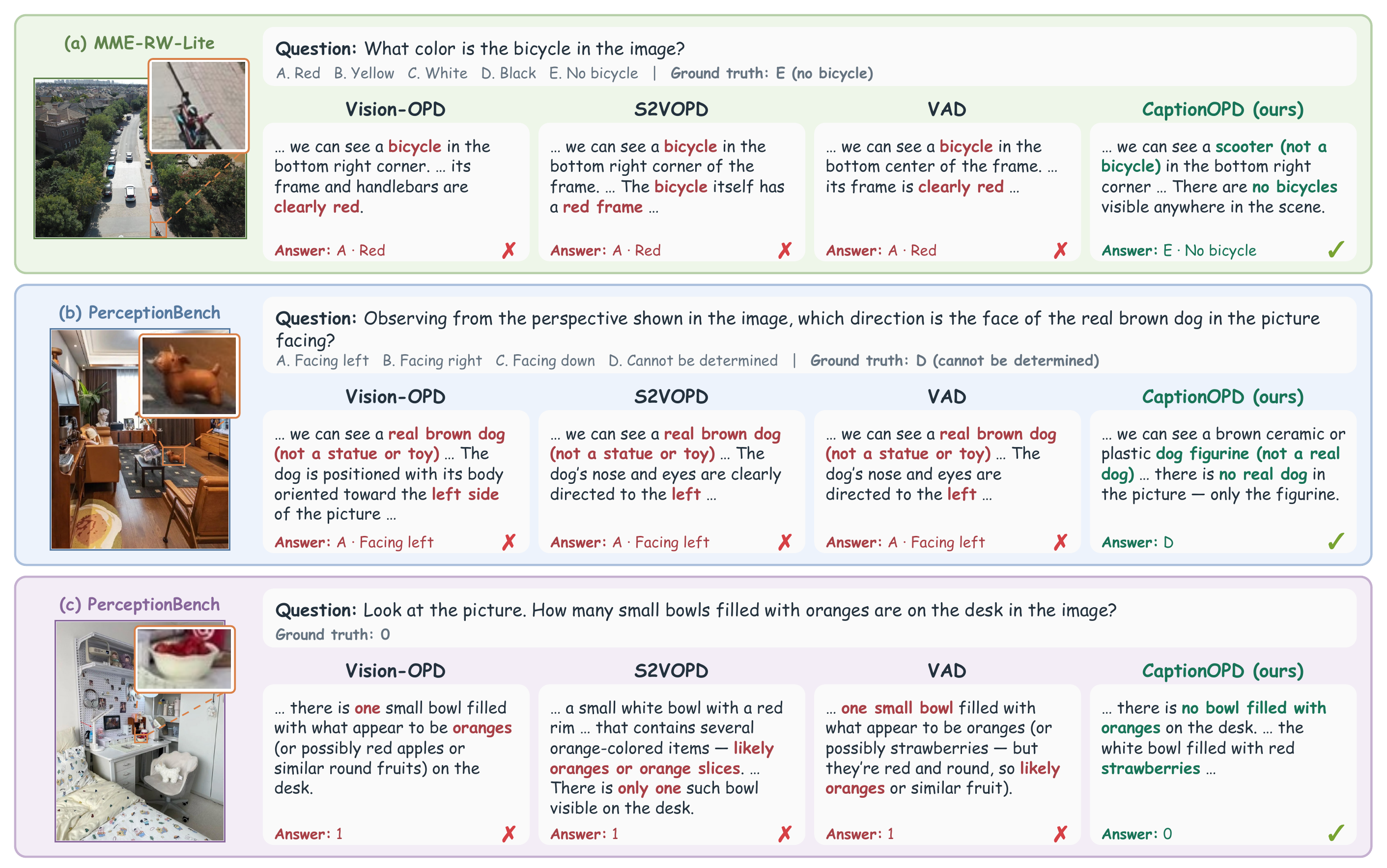}
    \caption{\textbf{Qualitative comparison on questions with false premises.}
    Each question presupposes an absent target:
    (a) a scooter rather than a bicycle,
    (b) a dog figurine rather than a real dog,
    and (c) a bowl of strawberries rather than oranges.
    QA-trained baselines (Vision-OPD, S$^2$VOPD, and VAD) blindly follow the question's premise
    and hallucinate the queried object,
    whereas CaptionOPD faithfully identifies what is actually present and rejects the false premise.
    All models are based on Qwen3.5-9B. More examples are shown in Appendix~\ref{sec:app-presupposition}.}
    \label{fig:presupposition}
\end{figure*}

\subsection{Main Results}
\label{sec:main-results}
 
\Cref{tab:main_benchmarks} reports the results on fine-grained perception.
Learning only from images and region boxes, without any question--answer annotations,
CaptionOPD achieves the highest average score on both model scales,
improving over the base model by 7.61 (4B) and 10.49 (9B) points
and outperforming other baselines.
Notably, on Qwen3.5-4B, all five QA-based OPD baselines fall below the base model on TreeBench,
whereas CaptionOPD improves.
Because CaptionOPD needs no QA annotation, its training data can be extended with unlabeled images.
Appendix~\ref{sec:app-scale} adds unlabeled SA-1B regions to Vision-OPD-6K (24,000 image-region pairs in total)
and shows that the average of ZoomGainBench and ZoomBench on Qwen3.5-9B keeps rising,
from 63.54 with Vision-OPD-6K alone to 66.06 after 250 steps.

\subsection{Generalization to Broader Perception Tasks}
\label{sec:general}

To examine whether the gains on fine-grained perception come at the cost of broader abilities,
we further evaluate the 9B models on eight general perception benchmarks (\cref{tab:general_benchmarks}).
Several OPD baselines (Vision-OPD, OPD-V, and VA-OPD) fall below the base model on average,
and all five baselines degrade on MMStar.
In contrast, CaptionOPD achieves the highest overall average.
The advantage is most striking on PerceptionBench,
where all five QA-based baselines drop below the base model,
whereas CaptionOPD improves.
CaptionOPD also shows clear gains on visual pattern and chart reasoning tasks,
such as MMVP and CharXiv-RQ.


\subsection{Describing Does Not Presuppose} 
\label{sec:presupposition}

A specific question inherently presupposes its target: for instance,
asking for the color of a bicycle assumes that a bicycle is present.
In Vision-OPD-6K, every question queries an existing object inside the region.
Consequently, training on such a dataset, where the queried target is always present,
conditions the student to blindly accept the question's premise.
In contrast, our caption prompts name no specific target
and simply ask the model to describe what is actually visible.
As shown in \cref{fig:presupposition} (and Appendix~\ref{sec:app-presupposition}),
this distinction is clear when models are asked about objects absent from the image.
Whereas QA-trained baselines (Vision-OPD, S$^2$VOPD, and VAD) conform to the false premise
and hallucinate attributes for the queried target,
CaptionOPD first identifies the actual object and then rejects the invalid query.


\subsection{Ablation Studies}
\label{sec:ablations}

\begin{table*}[t]
\centering
\caption{\textbf{Ablation of caption handoff and size-aware allocation.}
\textit{Uniform} assigns two general-caption rollouts to every region;
\textit{size-aware} assigns one, two, and three to small, medium, and large regions (\cref{sec:method-prompts}).
All models use Qwen3.5-9B.}
\label{tab:handoff_ablation}
\begingroup
\newcommand{\hoyes}{\textcolor{green!70!black}{\ding{51}}}
\newcommand{\hono}{\textcolor{black!60}{\ding{55}}}
\setlength{\tabcolsep}{3pt}  
\small
\resizebox{\textwidth}{!}{%
\renewcommand{\arraystretch}{1}
    \setlength{\aboverulesep}{1pt}
    \setlength{\belowrulesep}{1pt}
\begin{tabular}{@{}l|cc|*{8}{c}|c@{}}
\toprule
Method & Handoff & Allocation & \begin{tabular}[c]{@{}c@{}}ZoomGain\\Bench\end{tabular} & \begin{tabular}[c]{@{}c@{}}Zoom\\Bench\end{tabular} & V$^*$Bench & \begin{tabular}[c]{@{}c@{}}HR-Bench\\4K\end{tabular} & \begin{tabular}[c]{@{}c@{}}HR-Bench\\8K\end{tabular} & \begin{tabular}[c]{@{}c@{}}MME-RW\\Lite\end{tabular} & \begin{tabular}[c]{@{}c@{}}Tree\\Bench\end{tabular} & \begin{tabular}[c]{@{}c@{}}Visual\\Probe\end{tabular} & Avg. \\
\midrule
No handoff & \hono & uniform & 61.26 & 59.88 & 92.67 & 85.75 & 83.75 & 63.11 & 51.60 & 58.06 & 69.51 \\
+ Caption handoff & \hoyes & uniform & \textbf{66.00} & \textbf{61.66} & 93.72 & 85.75 & 84.88 & 63.26 & 52.35 & 59.81 & 70.93 \\
+ Size-aware allocation (\textbf{CaptionOPD}) & \hoyes & size-aware & \textbf{66.00} & 61.07 & \textbf{94.24} & \textbf{86.88} & \textbf{85.88} & \textbf{63.42} & \textbf{53.58} & \textbf{61.55} & \textbf{71.58} \\
\bottomrule
\end{tabular}%
}
\endgroup
\end{table*}

\begin{table*}[t]
\centering
\caption{\textbf{Ablation of the prompt pool.}
Each row changes only the prompts assigned to the 8 rollouts of a region.
\textit{Single} repeats one focused template, sampled per region, for all 8 rollouts;
\textit{Diverse} assigns distinct focused templates;
\textit{General only} repeats general caption prompt for all 8 rollouts.
General-caption rollouts use caption handoff as in the full method.
All models use Qwen3.5-9B.}
\label{tab:prompt_ablation}
\begingroup
\newcommand{\pdyes}{\textcolor{green!70!black}{\ding{51}}}
\newcommand{\pdno}{\textcolor{black!60}{\ding{55}}}
\setlength{\tabcolsep}{3pt}
\small
\resizebox{\textwidth}{!}{%
\renewcommand{\arraystretch}{1}
    \setlength{\aboverulesep}{1pt}
    \setlength{\belowrulesep}{1pt}
\begin{tabular}{@{}l|cc|*{8}{c}|c@{}}
\toprule
Prompts & \begin{tabular}[c]{@{}c@{}}General\\caption\end{tabular} & \begin{tabular}[c]{@{}c@{}}Focused\\caption\end{tabular} & \begin{tabular}[c]{@{}c@{}}ZoomGain\\Bench\end{tabular} & \begin{tabular}[c]{@{}c@{}}Zoom\\Bench\end{tabular} & V$^*$Bench & \begin{tabular}[c]{@{}c@{}}HR-Bench\\4K\end{tabular} & \begin{tabular}[c]{@{}c@{}}HR-Bench\\8K\end{tabular} & \begin{tabular}[c]{@{}c@{}}MME-RW\\Lite\end{tabular} & \begin{tabular}[c]{@{}c@{}}Tree\\Bench\end{tabular} & \begin{tabular}[c]{@{}c@{}}Visual\\Probe\end{tabular} & Avg. \\
\midrule
Single focused & \pdno & single & 58.82 & 59.05 & \textbf{94.24} & 86.00 & 83.75 & 62.69 & 49.38 & 56.31 & 68.78 \\
Diverse focused & \pdno & diverse & 60.94 & 60.12 & \textbf{94.24} & 84.50 & 85.00 & 63.31 & 50.37 & 57.09 & 69.45 \\
General only & \pdyes & \pdno & \textbf{67.87} & \textbf{61.66} & 92.67 & 85.75 & 83.12 & 61.59 & 50.37 & 58.25 & 70.16 \\
\midrule
General + diverse focused (\textbf{CaptionOPD}) & \pdyes & diverse & 66.00 & 61.07 & \textbf{94.24} & \textbf{86.88} & \textbf{85.88} & \textbf{63.42} & \textbf{53.58} & \textbf{61.55} & \textbf{71.58} \\
\bottomrule
\end{tabular}%
}
\endgroup
\end{table*}
\paragraph{Caption Handoff and Rollout Allocation.}
\Cref{tab:handoff_ablation} ablates the impact of caption handoff and size-aware rollout allocation on Qwen3.5-9B.
Handoff raises the average from 69.51 to 70.93,
improving seven of the eight benchmarks.
At test time, every benchmark provides only the unmarked image,
so the model must find the target from the question text alone.
This is the state that handoff trains:
after the first sentence names the target, the rest of the caption is supervised on the clean image.
The largest gain appears on ZoomGainBench (+4.74),
whose questions often refer to a small target through a long textual description.
Appendix~\ref{sec:app-binding} shows that without handoff the learned perception is bound to the red box:
removing the box at test time costs the no-handoff model 11.42 points on ZoomGainBench, twice the drop of CaptionOPD.
Allocating general-caption rollouts by region size (last row)
further improves the average,
with the largest gains on VisualProbe, TreeBench, and HR-Bench 4K/8K.

\paragraph{Prompt Diversity.} 
\Cref{tab:prompt_ablation} varies the prompts assigned to the rollouts of a region.
Diversity helps at two levels.
First, among focused prompts, assigning distinct templates to the rollouts of a region~(\textit{Diverse focused})
outperforms repeating a single template~(\textit{Single focused}),
improving six of the eight benchmarks:
when every rollout asks about a different aspect,
the same crop-view teacher supervises the student on more of the region.
Second, general and focused captions are complementary.
General captions excel on ZoomGainBench and ZoomBench,
while focused captions excel on V*Bench and MME-RW-Lite.
Mixing the two achieves the best average,
with the largest gains on TreeBench and VisualProbe.

\section{Related Work}

\paragraph{Fine-Grained Visual Perception in VLMs}

Fine-grained visual perception requires VLMs to locate and interpret subtle details in complex, high-resolution scenes~(\cite{shen2026zoomgain,wei2026zooming,wu2023vstar,wang2024hrbench,zhang2024mmerealworld,wang2025treebench,lai2025mini-o3}).
While early fixed-resolution models~(\cite{llava,Li2023BLIP2BL,InstructBLIP,Qwen-VL}) readily lose fine details and later dynamic-resolution architectures~(\cite{Qwen2-VL,Qwen2.5-VL,liu2024llavanext,llava-ov,internvl2-5}) partially alleviate this issue, recognizing tiny objects in cluttered scenes remains challenging.
To supply clearer local views at test time, V*~(\cite{wu2023vstar}) performs guided visual search, and related methods adopt training-free decomposition, zooming, or retrieval~(\cite{wang2024hrbench,shen2025zoomeye,rap,ImageRAG}).
Inspired by OpenAI o3~(\cite{openai2025systemcardo3o4}), recent works further use reinforcement learning to train models to autonomously invoke crop-and-zoom tools during inference~(\cite{zheng2026deepeyes,hong2026deepeyesv2,zhang2025thymethinkimages,su2026pixel,SenseNova-MARS}), though multi-turn tool calling inevitably increases latency.
To avoid test-time cropping overhead, ZwZ~(\cite{wei2026zooming}) distills supervision generated from zoomed crops into a model that answers directly on the full image.

\paragraph{On-Policy Distillation for VLMs}

On-policy distillation (OPD) provides token-level teacher supervision along student-sampled trajectories, mitigating the distribution shift of offline distillation with high sample efficiency~(\cite{agarwal2024onpolicy,minillm,lu2025onpolicydistillation}).
Subsequent works extend OPD to self-distillation using environment feedback or privileged context~(\cite{Hbotter2026ReinforcementLV,zhao2026opsd}), and adapt it to multimodal and visual mathematical reasoning~(\cite{Bousselham2025VOLDRT,Hou2026UniOPDUO,Ye2026EditDI,Tong2026MMOPDTO,Xue2026DistillWT,liu2026visualadvantageonpolicydistillationvisionlanguage}).
For fine-grained visual perception, Vision-OPD~(\cite{yuan2026visionopd}) distills a crop-conditioned teacher into a full-image student, while S$^2$VOPD~(\cite{li2026selfsupervisedvisualonpolicydistillation}) creates a view-quality gap by pairing degraded student views with clean teacher views.
Furthermore, OPD-V~(\cite{opdv2026}) contrasts teacher predictions across zoomed-in and masked views, and VAD~(\cite{zhang2026vad}) compares predictions with and without visual evidence to retain visually grounded corrections.

\section{Conclusion}

In this paper, we presented \textbf{CaptionOPD}, a on-policy distillation framework
that replaces region-specific question answering with multi-aspect regional captioning.
By pairing each region's rollouts with diverse, image-agnostic templates,
CaptionOPD adopts a crop-view self-teacher to supervise the full-image student across a rich set of visual details.
Furthermore, \emph{caption handoff} anchors general captions with a brief box-conditioned prefix
and distills the remaining description on the unmarked full image,
effectively transferring holistic crop knowledge without making the student rely on bounding boxes during inference.
Under the same regions and rollout budget,
CaptionOPD consistently outperforms QA-based OPD methods on both fine-grained and general perception benchmarks,
while enabling the model to reliably reject false premises when queried about absent targets.
Requiring only images and region boxes without any QA annotations or larger teacher models,
CaptionOPD offers a practical path toward scaling fine-grained visual distillation.

\bibliography{main}
\bibliographystyle{antgroup}

\clearpage
\appendix
\numberwithin{equation}{section}
\renewcommand{\thefigure}{S\arabic{figure}}
\renewcommand{\thetable}{S\arabic{table}}
\setcounter{figure}{0}
\setcounter{table}{0}
\section{Scaling CaptionOPD with Unlabeled Images}
\label{sec:app-scale}

\begin{wrapfigure}{R}{0.5\textwidth}
    \centering
    \includegraphics[width=\linewidth]{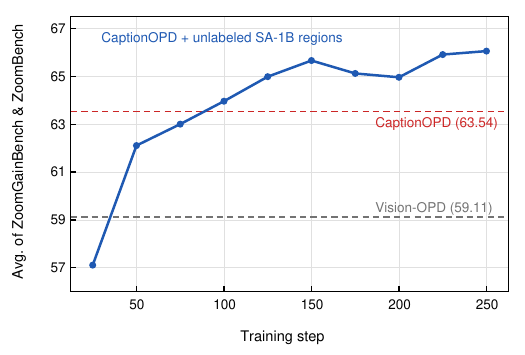}
    \caption{\textbf{Scaling with unlabeled images} (Qwen3.5-9B).
    Average score of ZoomGainBench and ZoomBench versus training step
    when Vision-OPD-6K is extended with unlabeled SA-1B regions to 24,000 image-region pairs.
    Dashed lines: CaptionOPD and Vision-OPD trained on Vision-OPD-6K (\cref{tab:main_benchmarks}).}
    \label{fig:app-scale}
\end{wrapfigure}

This section supports the claim in \cref{sec:intro} that QA-based OPD is hard to scale to unlabeled images,
whereas CaptionOPD is not.
CaptionOPD needs only an image and a region bounding-box,
so its training data can easily scale without any requirements regarding heavy question-answer annotation.
We evaluate this through extending the Vision-OPD-6K training set with unlabeled images.

\paragraph{Data.}
We use images from SA-1B~(\cite{kirillov2023segment}), the image source of Vision-OPD-6K,
and obtain region boxes from WeDetect~(\cite{fu2025wedetect}).
We keep one box per image and match the distribution of box areas to that of Vision-OPD-6K.
Images that appear in ZoomGainBench or Vision-OPD-6K are excluded.
Together with Vision-OPD-6K, this gives 24,000 image-region pairs in total.
Each region receives $K=8$ prompts as in \cref{sec:exp-settings},
yielding 192,000 training instances.

\paragraph{Training.}
We train Qwen3.5-9B with the same configuration as in \cref{sec:exp-settings}
(learning rate $2\times 10^{-6}$, global batch size 768),
for one epoch, which spents exactly 250 steps.
The teacher is updated by EMA during the first 65 steps and fixed afterwards.

\paragraph{Results.}
\Cref{fig:app-scale} reports the average score of ZoomGainBench and ZoomBench along training.
These two benchmarks closely match the setting that fine-grained visual OPD targets.
Each question depends on a tiny region of a high-resolution image with complex scene,
and the answer becomes easy to perceive once the region is zoomed in.
This is exactly the gap between the teacher seeing the zoomed region,
and the student seeing  the full image.
Thus, these two benchmarks directly measure how much of the teacher's zoomed perception
the student has learned, and we use them to track the effect of scaling.
The score rises from 57.11~(step 25) to 66.06~(step 250).
It exceeds Vision-OPD (59.11) at step 50
and CaptionOPD trained on Vision-OPD-6K (63.54) at step 100.
These results show that CaptionOPD keeps improving as more unlabeled images are introduced,
requiring no collecting any annotated QA pairs.

\section{Caption Handoff Removes Red-Box Binding}
\label{sec:app-binding}

\begin{wraptable}{R}{0.5\textwidth}
\centering
\caption{\textbf{Red-box binding} (Qwen3.5-9B).
\textit{w/o box}: original benchmark images (scores from \cref{tab:handoff_ablation});
\textit{w/ box}: the evidence region marked by a red box with the training focus instruction.
$\Delta$: change when the box is removed.}
\label{tab:app-binding}
\setlength{\tabcolsep}{3pt}
\small
\resizebox{\linewidth}{!}{%
\begin{tabular}{@{}l|ccc|ccc@{}}
\toprule
\multirow{2}{*}{Method} & \multicolumn{3}{c|}{ZoomGainBench} & \multicolumn{3}{c}{ZoomBench} \\
 & w/ box & w/o box & $\Delta$ & w/ box & w/o box & $\Delta$ \\
\midrule
No handoff & \textbf{72.68} & 61.26 & $-$11.42 & \textbf{64.85} & 59.88 & $-$4.97 \\
CaptionOPD & 71.75 & \textbf{66.00} & $-$\textbf{5.75} & 63.43 & \textbf{61.07} & $-$\textbf{2.36} \\
\bottomrule
\end{tabular}%
}
\end{wraptable}

This section extends the caption handoff ablation in \cref{sec:ablations} (\cref{tab:handoff_ablation}),
with the method described in \cref{sec:method-handoff}.
Without caption handoff, every caption is trained on the image with the red box drawn on it.
In this section, we test whether the perception capability learned in this way is bound to the red box itself.
We evaluate the no-handoff model and CaptionOPD on ZoomGainBench and ZoomBench
under two view settings:
(1)~the original image (\textit{w/o box}),
and (2)~the same image with the evidence region marked by a red box
together with the focus instruction, which is the same as the configuration used in training (\textit{w/ box}).

\Cref{tab:app-binding} shows that the two models perform similarly with the box drawing.
Without the box, however, the no-handoff model drops by 11.42 points on ZoomGainBench,
whose questions requires the model to locate the target through a textual description,
and by 4.97 points on ZoomBench.
CaptionOPD drops by only 5.75 and 2.36 points,
and outperforms the no-handoff model on two benchmarks.
Without handoff, the learned perception capability is bound to the red box:
the model perceives the target well only when the box marks it.
Caption handoff trains the caption on the unmarked image and use its first sentence to point the target out,
so the student learns to find the region from the description itself.

\section{VAD Support and Refutation Budgets}
\label{sec:app-vad}

We detail the branch allocation used to construct $\delta_t$ in \cref{eq:vad-target},
following VAD~(\cite{zhang2026vad}).
The student, crop teacher, and degraded-crop teacher distributions share the same coordinates:
the student's top-$K$ token candidates and a tail bucket containing the remaining probability mass.
The teacher probabilities are gathered on the same token identifiers before applying $\phi$.

The positive and negative coordinates of $u_t$ represent relative support and refutation.
VAD separates them and measures their agreement with the teacher--student difference $r_t$:
\begin{equation}
    \begin{aligned}
        u_t^+&=[u_t]_+,\qquad u_t^-=[u_t]_-,\\
        s_t^\pm&=[\langle r_t,u_t^\pm\rangle]_+.
    \end{aligned}
    \label{eq:vad-branch-scores}
\end{equation}
Here $[a]_+=\max(a,0)$ and $[a]_-=\min(a,0)$ act coordinatewise on vectors;
the refutation branch retains its negative sign.
With $Z_t=s_t^++s_t^-+\epsilon$, the allocation weights are
\begin{equation}
    \omega_t^+=\min\!\left(\frac{s_t^+}{Z_t},\tau_+\right),
    \qquad \omega_t^-=\frac{s_t^-}{Z_t}.
    \label{eq:vad-branches}
\end{equation}
Only the support share is capped by $\tau_+\in[0,1]$;
the removed share is not reassigned to the refutation branch.

The projected component in \cref{eq:vad-projection} supplies the budget
$B_t=\|r_t^{\mathrm{vis}}\|_2$.
The final shift combines the normalized branches using their allocated budgets:
\begin{equation}
    \delta_t=B_t\!\left(
    \omega_t^+\frac{u_t^+}{\|u_t^+\|_2+\epsilon}
    +\omega_t^-\frac{u_t^-}{\|u_t^-\|_2+\epsilon}\right).
    \label{eq:vad-budgeted-shift}
\end{equation}
Thus, $\delta_t$ retains the projection-based budget while allowing separate control
of support and refutation.
When the one-sided projection gives $\beta_t=0$, the budget and shift are both zero.
The target in \cref{eq:vad-target} uses the clipped shift,
whereas the regularizer weight in \cref{eq:vad-objective} uses its norm before clipping.

\section{Caption Handoff Details}
\label{sec:app-handoff}

We detail two implementation choices of caption handoff (\cref{sec:method-handoff}).

\paragraph{First-sentence boundary.}
The first sentence of a general caption ends at the first period, exclamation mark,
or question mark that is followed by whitespace.
This character position is mapped to the corresponding token position,
which gives the first-sentence length $L^{\mathrm{fs}}$ in \cref{eq:handoff-loss}.
The first $L^{\mathrm{fs}}$ tokens remain in the student's prefix as context,
and only the subsequent tokens are supervised.

\paragraph{First sentences that mention the box.}
The student writes its caption with the box visible,
so the first sentence occasionally refers to the box itself
(\eg, ``The object inside the red box is \ldots'')
instead of naming the target by its appearance.
Such a sentence cannot serve as a textual anchor on the unmarked image,
and the rest of the caption may continue to refer to the box.
We therefore detect box-referring phrases (\eg, ``red box'', ``bounding box'')
in the first sentence and exclude the whole caption from the loss
by setting all entries of its supervision mask $\mathbf m$ to zero.
The caption then contributes to neither the numerator nor the denominator
of \cref{eq:caption-opd-objective}.

\section{More Examples of Questions about Absent Targets}
\label{sec:app-presupposition}

\newcommand{\nullimg}[1]{\raisebox{\dimexpr-\height+1.6ex\relax}{\includegraphics[width=\linewidth]{#1}}}
\newcommand{\nullwrong}[1]{\par\vspace{2pt}\hfill\textbf{#1}~\textcolor{red!75!black}{\ding{55}}}
\newcommand{\nullright}[1]{\par\vspace{2pt}\hfill\textbf{#1}~\textcolor{green!50!black}{\ding{51}}}

\begin{figure*}[t]
    \centering
    \scriptsize
    \setlength{\tabcolsep}{3pt}
    \renewcommand{\arraystretch}{1.15}
    \begin{tabular}{@{}p{0.1\textwidth}p{0.155\textwidth}|p{0.16\textwidth}p{0.16\textwidth}p{0.16\textwidth}p{0.16\textwidth}@{}}
        \toprule
        \multicolumn{2}{@{}c|}{\textbf{Image and question}} & \multicolumn{1}{c}{\textbf{Vision-OPD}}
        & \multicolumn{1}{c}{\textbf{S2VOPD}} & \multicolumn{1}{c}{\textbf{VAD}}
        & \multicolumn{1}{c@{}}{\textbf{CaptionOPD (ours)}} \\
        \midrule
        \nullimg{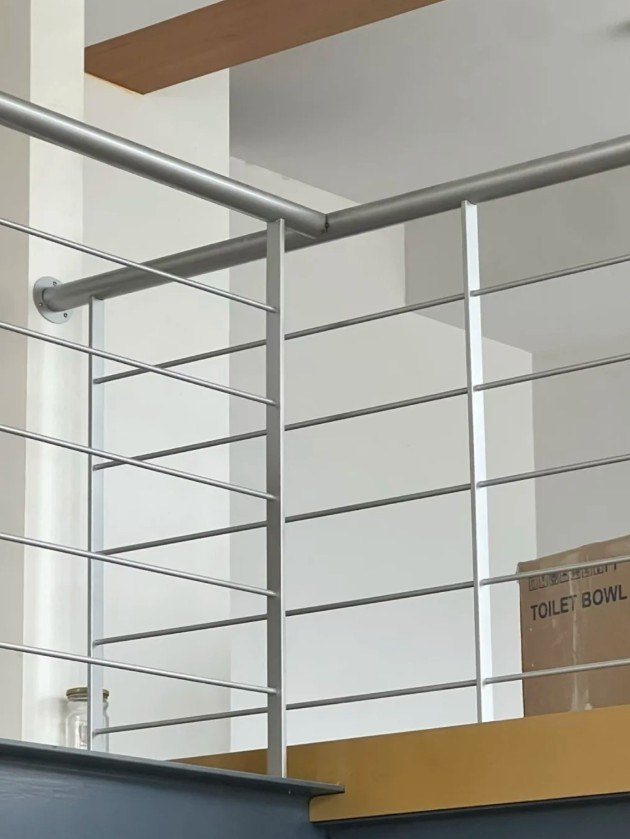} &
        \textbf{Q:} Look at the picture.
        How many steps are visible in the scene?
        \newline\textbf{GT:} 0 &
        \ldots\ we can see a staircase with a modern metal railing system.
        The steps are partially visible at the bottom of the frame. \ldots\
        only \textbf{three} are clearly visible in the image.
        \nullwrong{3} &
        \ldots\ we can see a staircase with a modern metal railing system. \ldots\
        Therefore, \textbf{4 steps} are visibly distinguishable in the scene.
        \nullwrong{4} &
        \ldots\ there is a visible section of what appears to be a staircase or landing area. \ldots\
        \textbf{only one tread is clearly visible} --- the mustard-yellow one.
        \nullwrong{1} &
        Actually, \textbf{there are no steps visible} in the scene.
        The image shows a modern metal railing system \ldots\
        above a mustard-yellow surface (possibly a ledge or countertop). \ldots
        \nullright{0} \\
        \midrule
        \nullimg{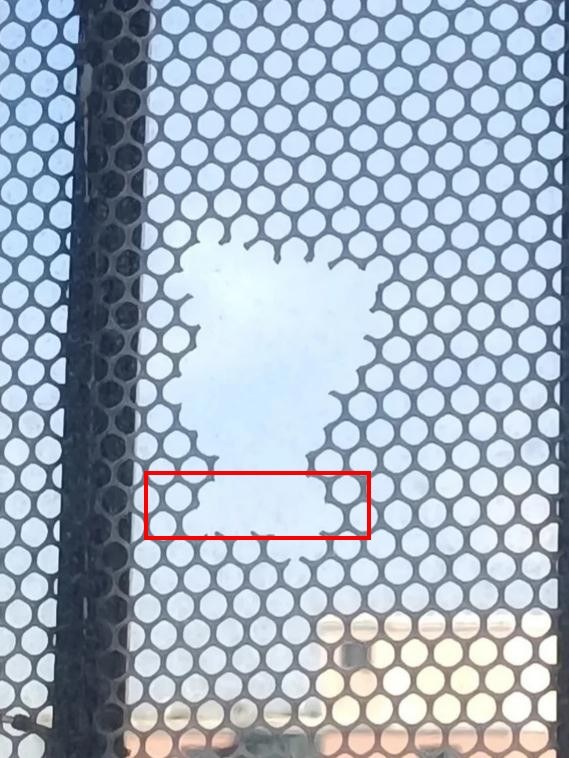} &
        \textbf{Q:} Based on the information in the image,
        how many birds are there in the sky in total?
        A.~0 birds B.~1 bird C.~2 birds D.~3 birds
        \newline\textbf{GT:} A &
        \ldots\ Through this hole, we can observe the sky and what appears to be
        \textbf{two distinct bird silhouettes} --- one on the left side of the hole
        and one on the right side.
        \nullwrong{C} &
        \ldots\ Through that hole, we can see the sky and what appears to be
        \textbf{two birds} flying --- one on the left side of the hole
        and one on the right side.
        \nullwrong{C} &
        \ldots\ we can see the sky and what appears to be \textbf{two birds} flying \ldots\
        These are distinct shapes that resemble birds in flight.
        \nullwrong{C} &
        \ldots\ we see a perforated metal sheet with circular holes.
        The background shows a pale sky and some distant ground or structures ---
        but \textbf{no birds} are visible anywhere in the scene.
        \nullright{A} \\
        \midrule
        \nullimg{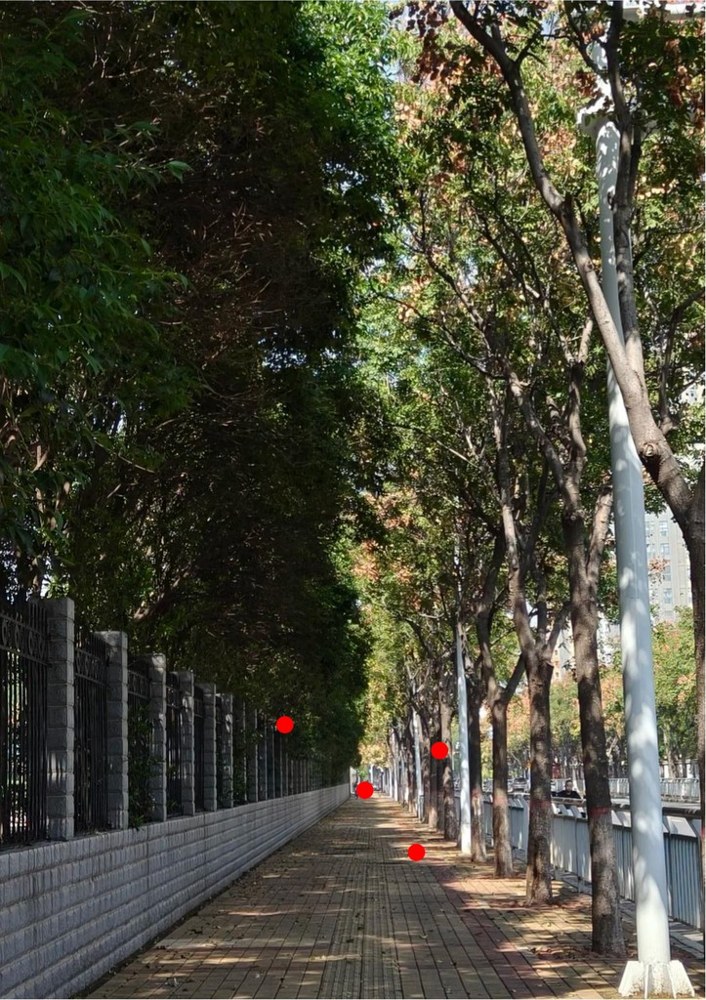} &
        \textbf{Q:} How many small animals are there in the tree in the picture?
        A.~0 animals B.~1 animal C.~2 animals D.~3 animals
        \newline\textbf{GT:} A &
        \ldots\ In the upper right portion of the image, there is a
        \textbf{small white animal} --- likely a squirrel or similar rodent ---
        perched on a branch. \ldots
        \nullwrong{B} &
        \ldots\ Upon close inspection of the tree (especially near the top branches),
        there is \textbf{one small animal} --- likely a squirrel or similar rodent ---
        visible in the foliage.
        \nullwrong{B} &
        \ldots\ nestled among the branches and leaves, there is a
        \textbf{small white animal} --- likely a squirrel or similar rodent ---
        visible near the top of the tree.
        \nullwrong{B} &
        \ldots\ The right side has trees with some brown leaves, but again,
        \textbf{no birds, squirrels, insects, or other small animals} are clearly visible. \ldots
        \nullright{A} \\
        \bottomrule
    \end{tabular}
    \caption{\textbf{More questions about absent targets (1/2).}
    There are no steps, only a railing above a ledge;
    no birds, only sky behind a perforated sheet;
    and no animals in the trees.
    All models are based on Qwen3.5-9B; responses are abridged.}
    \label{fig:app-presupposition-1}
\end{figure*}

\begin{figure*}[t]
    \centering
    \scriptsize
    \setlength{\tabcolsep}{3pt}
    \renewcommand{\arraystretch}{1.15}
    \begin{tabular}{@{}p{0.1\textwidth}p{0.155\textwidth}|p{0.16\textwidth}p{0.16\textwidth}p{0.16\textwidth}p{0.16\textwidth}@{}}
        \toprule
        \multicolumn{2}{@{}c|}{\textbf{Image and question}} & \multicolumn{1}{c}{\textbf{Vision-OPD}}
        & \multicolumn{1}{c}{\textbf{S2VOPD}} & \multicolumn{1}{c}{\textbf{VAD}}
        & \multicolumn{1}{c@{}}{\textbf{CaptionOPD (ours)}} \\
        \midrule
        \nullimg{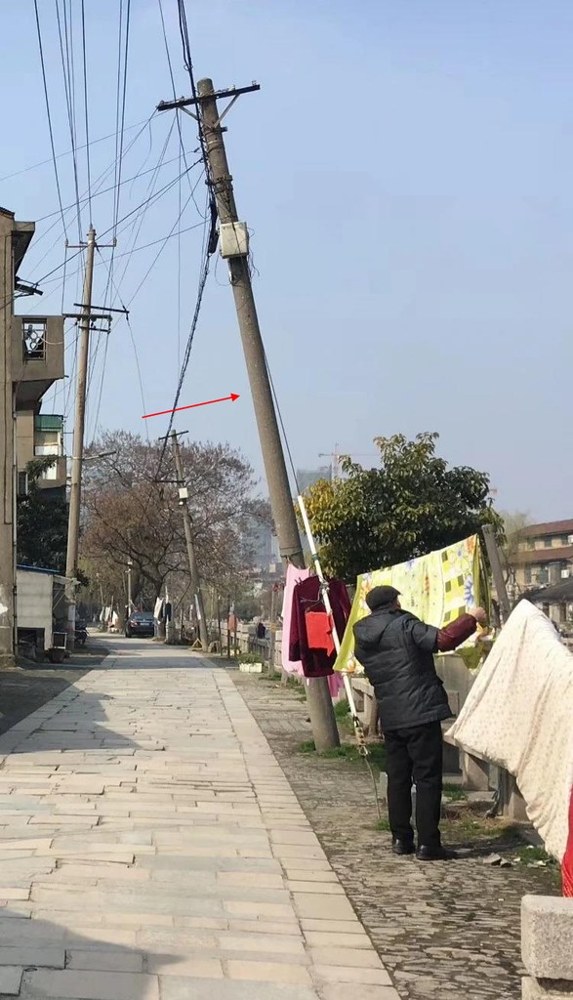} &
        \textbf{Q:} In the image, how many animal patterns are printed on the bedsheet
        closest to the person wearing a black cotton-padded jacket?
        \newline\textbf{GT:} 0 &
        \ldots\ This bedsheet has a light background with a repeating pattern of small,
        dark shapes that resemble \textbf{animal silhouettes} ---
        likely dogs or similar quadrupeds \ldots
        \nullwrong{12} &
        \ldots\ these shapes are clearly \textbf{animal patterns} ---
        specifically, they appear to be \textbf{small black cats}
        or similar feline figures \ldots
        \nullwrong{4} &
        \ldots\ the pattern consists of \textbf{small dogs} ---
        specifically, they appear to be stylized or cartoonish dogs \ldots
        \nullwrong{7} &
        \ldots\ The bedsheet closest to this person is the one with a
        \textbf{yellow background and floral patterns} \ldots\
        The visible patterns are floral/geometric, \textbf{not animal-based}.
        \nullright{0} \\
        \midrule
        \nullimg{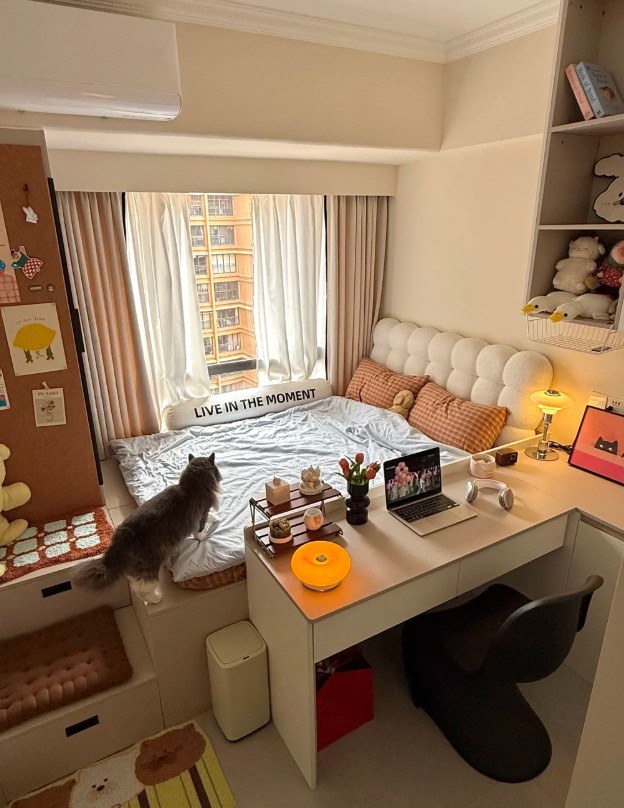} &
        \textbf{Q:} Observe the image.
        How many mice can be seen on the table in total?
        \newline\textbf{GT:} 0 &
        \ldots\ on its keyboard area, there is a \textbf{small white object}
        that resembles a computer mouse. \ldots
        \nullwrong{1} &
        \ldots\ To the right of the laptop, there is a
        \textbf{small white object that looks like a computer mouse} \ldots
        \nullwrong{1} &
        \ldots\ near the laptop and headphones, there is a
        \textbf{small white object that resembles a computer mouse} \ldots
        \nullwrong{1} &
        \ldots\ there are \textbf{no computer mice visible} on the table.
        The pink plush item might be mistaken for a mouse due to its shape,
        but it appears to be a decorative stuffed animal \ldots
        \nullright{0} \\
        \midrule
        \nullimg{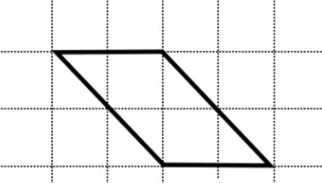} &
        \textbf{Q:} Observe the four sides of the parallelogram in the image.
        Among these four sides, the one drawn with a dashed line,
        from the observer's perspective looking at the image, is which side?
        A.~Top side B.~Bottom side C.~Left side D.~Right side
        E.~Cannot be determined
        \newline\textbf{GT:} E &
        \ldots\ the \textbf{left slanted side} of the parallelogram is drawn with a
        \textbf{dotted line}, while the other three sides (top, bottom, right) are solid.
        \nullwrong{C} &
        \ldots\ except for \textbf{one side} --- the \textbf{left slanted side} ---
        which is drawn with a \textbf{dashed line}.
        \nullwrong{C} &
        \ldots\ except for one side --- the \textbf{left slanted side} ---
        which is also drawn with a \textbf{dashed (or dotted) line}.
        \nullwrong{C} &
        \ldots\ There are dotted grid lines in the background,
        but \textbf{none of the parallelogram's sides are drawn with dashed lines}.
        \nullright{E} \\
        \bottomrule
    \end{tabular}
    \caption{\textbf{More questions about absent targets (2/2).}
    The bedsheet nearest the person has floral patterns,
    the desk holds no computer mouse,
    and all four sides of the parallelogram are solid;
    only the background grid is dotted.
    All models are based on Qwen3.5-9B; responses are abridged.}
    \label{fig:app-presupposition-2}
\end{figure*}

\Cref{fig:app-presupposition-1,fig:app-presupposition-2} show six more examples
in the setting of \cref{sec:presupposition}.
All six are all five question-driven OPD methods answer incorrectly whereas CaptionOPD answers in correct.
The questions ask about steps, birds, small animals, animal patterns, computer mice and a dashed side,
none of which appears in the image.
Vision-OPD, S2VOPD and VAD report the queried objects, often with a specific count.
CaptionOPD describes what the image shows instead
and concludes that the queried object is absent.

\clearpage
\begingroup
\setlength{\tabcolsep}{4pt}
\section{Caption Prompt Pool}
\label{sec:app-prompts}

This section lists the exact prompts used by CaptionOPD (\cref{sec:method-prompts}).
The prompt pool contains one general caption prompt and ten focused prompts.
The student and the crop-view teacher receive the same prompt text for every rollout.

\paragraph{General caption prompt.}
The general caption prompt is the box instruction followed by the description request:

\begin{quote}
\small
\textit{Look at the region inside the red box. Do not mention the red box itself.
Describe the content of this region in 3-5 declarative sentences, densely packed with concrete visual facts:
name every distinct object with its color, material, and state;
give an exact count whenever objects are countable (e.g., `three bolts', `twelve posters') - count precisely and never approximate a count;
and transcribe any readable text or numbers verbatim in quotation marks.
State the relative positions of the main objects.
Write each sentence as a direct factual statement about the scene itself - do not refer to `the image', `the crop', `the frame', or the act of viewing,
and do not use lists, headings, or markdown.
No filler, no speculation, and no meta-commentary about resolution or blur:
if a piece of text is genuinely unreadable, simply omit it rather than guessing or announcing that you cannot read it.}
\end{quote}

\noindent Under caption handoff (\cref{sec:method-handoff}),
the student's training input drops the first two sentences (the box instruction) and the red box,
while the teacher keeps the full prompt.

\paragraph{Focused caption prompts.}
Each focused prompt consists of three parts:
an aspect-specific request (\cref{tab:app-prompts}),
the box instruction
\textit{``Only focus on the objects inside the red bounding box in the image to answer this question.''},
and a response instruction.
Most prompts first ask for a short description of the object and then a one-line final answer:

\begin{quote}
\small
\textit{(a) First, in one short sentence, say what you see on the object itself (its parts, colors, markings)
that matters for this question; do not talk about the box or the image.
Then write the final answer on the last line, in this format: [format]}

\textit{(b) First, in one short sentence, say what small details you see on the object itself (small parts, colors, markings)
that matter for this question; do not talk about the box or the image.
Then write the final answer on the last line, in this format: [format]}

\textit{(c) Answer in one line, in this format: [format]}
\end{quote}

\noindent where [format] is the answer format in \cref{tab:app-prompts}.
None of the prompts names an image-specific object,
and the prompts that may have no valid target (attached items, counting, and text)
explicitly allow the answer \textit{none}.
For the spatial-layout prompt, the side is sampled once per region
from \{top, bottom, left side, right side, center\}.

\begin{table*}[t]
\centering
\caption{\textbf{Focused caption prompts.}
Each prompt is followed by the box instruction and the response instruction (a), (b), or (c) given in the text.}
\label{tab:app-prompts}
\small
\renewcommand{\arraystretch}{1.15}
\begin{tabular}{@{}p{0.11\textwidth}p{0.49\textwidth}p{0.3\textwidth}c@{}}
\toprule
Aspect & Request & Answer format & Resp. \\
\midrule
Part color & What is the main object? Name one large part of it, then one small detail on that part, and give the color of the small detail.
 & Object: [name]. Part: [large part]. Detail: [small detail]. Color: [color] & (a) \\
Detail color & What is the main object? Name the smallest part of it whose color you can still make out (for example a button, logo, buckle, handle, knob, or light), and give that color.
 & Object: [name]. Small part: [part]. Color: [color] & (a) \\
Attached items & What is the main object? Name one smaller thing attached to it, held by it, worn by it, or resting on it. If there is none, write Item: none.
 & Object: [name]. Item: [thing] & (b) \\
Category & What is the main object? Say what kind it is as specifically as the picture allows, such as its type, model, breed, or species.
 & Object: [name]. Kind: [specific kind] & (b) \\
Text & What is the main object? Copy the text, numbers, or symbols written on it exactly as they appear. If nothing is written on it, write Text: none.
 & Object: [name]. Text: ``[exact text]'' & (a) \\
Counting & What is the main object? Pick one kind of part or item on it that appears more than once, and count them. Write the count in digits, or ``more than 10''. If nothing repeats, write Part: none.
 & Object: [name]. Part: [repeated part]. Count: [number] & (c) \\
Texture & What is the main object, and what pattern or surface texture does it have?
 & Object: [name]. Pattern: [pattern or texture] & (b) \\
Material & What is the main object, and what is it mainly made of? If it is a person, give the material of their clothing.
 & Object: [name]. Material: [material] & (b) \\
Spatial layout & What is the main object? What is at the [side] of it, as seen in the picture?
 & Object: [name]. At the [side]: [what is there] & (a) \\
State & What is the main object, and what visible state is it in, such as its pose, its condition, or the direction it faces or points as seen in the picture?
 & Object: [name]. State: [a few words] & (a) \\
\bottomrule
\end{tabular}
\end{table*}

\endgroup
\clearpage
\section{Evaluation Prompts and LLM Judging}
\label{sec:app-eval}

\paragraph{Generation.}
All models are evaluated with the identical prompt for each benchmark,
with thinking mode disabled, greedy decoding, and a maximum of 4,096 output tokens.
Each benchmark keeps its own question format (\cref{tab:app-bench-prompts}).

\begin{table*}[t]
\centering
\caption{\textbf{Prompt templates of the evaluation benchmarks.}
\texttt{\{question\}} is the benchmark question, and options are listed one per line unless stated otherwise.
\textit{Rule}: the answer is first extracted by a strict rule (an option letter or a final ``Answer:'' line);
models responses the rule cannot parse are sent to the LLM judge.}
\label{tab:app-bench-prompts}
\small
\renewcommand{\arraystretch}{1.15}
\begin{tabular}{@{}p{0.15\textwidth}p{0.5\textwidth}p{0.29\textwidth}@{}}
\toprule
Benchmark & Prompt & Scoring \\
\midrule
ZoomGainBench & Multiple choice: \texttt{\{question\}} with options \texttt{A.}~\ldots;
counting: \texttt{\{question\}\textbackslash nPlease answer using Arabic numerals.};
OCR: \texttt{\{question\}\textbackslash nAnswer with only the transcription. Preserve the original capitalization, spacing, punctuation, and symbols.}
 & Judge extracts the final answer (\cref{fig:app-judge-zgb}); exact match for options and counts, ANLS-based credit for OCR \\
ZoomBench & \texttt{\{question\}} with options \texttt{A.}~\ldots, or a numeric question & Rule, then judge (\cref{fig:app-judge-vopd}) \\
V$^*$Bench & \texttt{\{question\}} with options \texttt{(A)}~\ldots & Rule, then judge (\cref{fig:app-judge-vopd}) \\
HR-Bench 4K/8K & \texttt{\{question\} Select from the following choices.\textbackslash n} followed by options \texttt{A.}~\ldots & Rule, then judge (\cref{fig:app-judge-vopd}) \\
MME-RW-Lite & \texttt{\{question\} The choices are listed below:\textbackslash n} followed by options \texttt{(A)}~\ldots, then
\texttt{\textbackslash nSelect the best answer to the above multiple-choice question based on the image.} & Rule, then judge (\cref{fig:app-judge-vopd}) \\
TreeBench & \texttt{\{question\}\textbackslash nOptions:\textbackslash n} followed by options \texttt{A.}~\ldots & Rule, then judge (\cref{fig:app-judge-vopd}) \\
VisualProbe & \texttt{\{question\}} (open-ended) & Judge (\cref{fig:app-judge-vp}) \\
\midrule
PerceptionBench & \texttt{\{question\}} as released, with all images before the text & Rule, then the official PerceptionBench judge prompt \\
MMVP & \texttt{\{question\} (a) \ldots\ (b) \ldots} & Rule, then judge (\cref{fig:app-judge-vopd}); pair accuracy \\
BLINK & Official prompt with options \texttt{(A)}~\ldots, all images before the text & Rule, then judge (\cref{fig:app-judge-vopd}) \\
CV-Bench & Official prompt with options \texttt{(A)}~\ldots & Rule, then judge (\cref{fig:app-judge-vopd}); source-balanced accuracy \\
CharXiv-DQ/RQ & Official question with its official response instructions & Official grading prompts of CharXiv~(\cite{wang2024charxiv}); truncated replies score 0 \\
SEED-Bench-2-Plus & \texttt{\{question\}} with options \texttt{A.}~\ldots & Rule, then judge (\cref{fig:app-judge-vopd}); truncated replies score 0 \\
MMStar & \texttt{\{question\}} with inline options \texttt{Options: A: \ldots, B: \ldots} as released & Rule, then judge (\cref{fig:app-judge-vopd}) \\
\bottomrule
\end{tabular}
\end{table*}

\paragraph{LLM judge.}
Models' responses that the strict rule cannot parse are judged by Qwen3.8-27B~(\cite{qwen3.8})
with thinking disabled and temperature 0.
Unless a benchmark provides its exclusive judge prompt,
we use the judge prompt of Vision-OPD~(\cite{yuan2026visionopd}) (\cref{fig:app-judge-vopd}).
VisualProbe uses its official judge prompt (\cref{fig:app-judge-vp}),
and PerceptionBench and CharXiv use their official grading prompts.
For ZoomGainBench, the judge only extracts the final answer from the response
(\cref{fig:app-judge-zgb}), and the extracted answer is then scored against the ground truth.
\textbf{To verify the feasibility of Qwen3.8-27B, we manually check 400 judge result records, and all judgements are correct.}

\begin{figure}[t]
\centering
\small
\fbox{\parbox{0.95\columnwidth}{\ttfamily\footnotesize
Your task is to judge whether the response expresses the same meaning as the answer of a question.\\
The question is: \{question\}\\
The answer is: \{gt\}\\
The response is: \{response\}\\
Please check and compare them and then judge. If the response is correct, your output should be Yes. Otherwise, your output should be No. Directly give me your output.}}
\caption{\textbf{Default judge prompt} (from Vision-OPD).}
\label{fig:app-judge-vopd}
\end{figure}

\begin{figure}[t]
\centering
\small
\fbox{\parbox{0.95\columnwidth}{\ttfamily\footnotesize
You read a model's reply to a question about an image and copy out the final answer the reply gives. You do not see the image, and you must not judge whether the answer is right.\\[2pt]
Question:\\
\{question\}\\[2pt]
Reply:\\
\{reply\}\\[2pt]
\{task\}\\[2pt]
If the reply gives no single final answer (it refuses, says it cannot tell, or ends undecided between different answers), output NONE.
Output only the result between <final> and </final>, for example <final>\{example\}</final>.}}
\par\vspace{4pt}
\fbox{\parbox{0.95\columnwidth}{\ttfamily\footnotesize
\textrm{\textbf{Multiple choice} (example: B):}
The question is multiple choice; its options are listed in the question. Give the letter of the option the reply finally chooses. If the reply gives its answer as the text of an option, or describes its answer instead of naming a letter, give the letter of the one option that matches that answer; if the answer matches no option or more than one option, output NONE.\\[2pt]
\textrm{\textbf{Counting} (example: 7):}
The question asks for a count. Give the reply's final count in Arabic numerals (digits only). If the reply writes the number in words, in any language (for example "three"), convert it to digits.\\[2pt]
\textrm{\textbf{OCR} (example: OPEN 24 HOURS):}
The question asks for text read from the image. Copy the reply's final transcription exactly as the reply writes it, character for character: keep its capitalization, spacing, punctuation, symbols and language. Do not correct, complete, translate or reformat it. Leave out only wrapping that is not part of the transcription, such as surrounding quotation marks, Markdown markers like ** or `, and lead-ins like "The text reads:". A reply that is nothing but a piece of text is itself the transcription, even if that text reads like a greeting or a sentence: copy all of it.}}
\caption{\textbf{ZoomGainBench answer-extraction prompt} (top) and the \texttt{\{task\}} text for each question type (bottom).}
\label{fig:app-judge-zgb}
\end{figure}

\begin{figure}[t]
\centering
\small
\fbox{\parbox{0.95\columnwidth}{\ttfamily\footnotesize
\textrm{\textbf{System:}}
You are an intelligent chatbot designed for evaluating the correctness of generative outputs for question-answer pairs.
Your task is to compare the predicted answer with the correct answer and determine if they match meaningfully. Here's how you can accomplish the task:\\
------\\
\#\#INSTRUCTIONS:\\
- Focus on the meaningful match between the predicted answer and the correct answer.\\
- Consider synonyms or paraphrases as valid matches.\\
- Evaluate the correctness of the prediction compared to the answer.\\[2pt]
\textrm{\textbf{User:}}
I will give you a question related to an image and the following text as inputs:\\
1. **Question Related to the Image**: \{question\}\\
2. **Ground Truth Answer**: \{ground\_truth\}\\
3. **Model Predicted Answer**: \{prediction\}\\
Your task is to evaluate the model's predicted answer against the ground truth answer, based on the context provided by the question related to the image. Consider the following criteria for evaluation:\\
- **Relevance**: Does the predicted answer directly address the question posed, considering the information provided by the given question?\\
- **Accuracy**: Compare the predicted answer to the ground truth answer. You need to evaluate from the following two perspectives:\\
(1) If the ground truth answer is open-ended, consider whether the prediction accurately reflects the information given in the ground truth without introducing factual inaccuracies. If it does, the prediction should be considered correct.\\
(2) If the ground truth answer is a definitive answer, strictly compare the model's prediction to the actual answer. Pay attention to unit conversions such as length and angle, etc. As long as the results are consistent, the model's prediction should be deemed correct.\\
**Output Format**:\\
Your response should include an integer score indicating the correctness of the prediction: 1 for correct and 0 for incorrect. Note that 1 means the model's prediction strictly aligns with the ground truth, while 0 means it does not.\\
The format should be "Score: 0 or 1"}}
\caption{\textbf{VisualProbe judge prompt} (official).}
\label{fig:app-judge-vp}
\end{figure}

\end{document}